\documentclass[12pt,a4paper]{article}
\usepackage{pslatex}
\usepackage{colortbl}
\usepackage[pdftex]{graphicx}
\usepackage[margin=1.1in]{geometry}
\usepackage{setspace}
\usepackage{lscape}
\usepackage{multirow}
\usepackage{ctable}
\usepackage{subfigure}
\usepackage{amsmath}
\usepackage{natbib}
\usepackage{color}
\usepackage{eurosym}
\usepackage{booktabs}
\usepackage[toc,page]{appendix}

\newcolumntype{C}[1]{>{\centering\let\\\tabularnewline}p{#1}}
\newcolumntype{R}[1]{>{\raggedleft\let\\\tabularnewline}p{#1}}
\newcolumntype{L}[1]{>{\raggedright\let\\\tabularnewline}p{#1}}

\usepackage{hyperref}
\hypersetup{colorlinks=true, %
            linkcolor=black, %
            citecolor=black, %
            filecolor=black, %
            urlcolor=black,  %
            pdftitle={Dail debates database}, %
            pdfauthor={Alexander Herzog and Slava Mikhaylov}, %
            pdfsubject={Dail debates database}, %
            pdftex}

\begin{document}

\title{Database of Parliamentary Speeches in Ireland, 1919--2013\thanks{The database is made available on the Harvard Dataverse at \url{http://dx.doi.org/10.7910/DVN/6MZN76}.}}

\author{
Alexander Herzog\\
  Clemson University \\
  aherzog@clemson.edu \\
\and 
Slava J. Mikhaylov \\
 University of Essex\\
  s.mikhaylov@essex.ac.uk
}

\maketitle

\begin{abstract}
We present a database of parliamentary debates that contains the complete record of parliamentary speeches from D{\'a}il {\'E}ireann, the lower house and principal chamber of the Irish parliament, from 1919 to 2013. In addition, the database contains background information on all TDs (Teachta D{\'a}la, members of parliament), such as their party affiliations, constituencies and office positions. The current version of the database includes close to 4.5 million speeches from 1,178 TDs. The speeches were downloaded from the official parliament website and further processed and parsed with a Python script. Background information on TDs was collected from the member database of the parliament website. Data on cabinet positions (ministers and junior ministers) was collected from the official website of the government. A record linkage algorithm and human coders were used to match TDs and ministers.

\vspace{0.5cm}


\end{abstract}

\thispagestyle{empty}

\newpage
\begin{spacing}{2}

\section{Introduction}

Almost all political decisions and political opinions are, in one way or another, expressed in written or spoken texts. Great leaders in history become famous for their ability to motivate the masses with their speeches; parties publish policy programmes before elections in order to provide information about their policy objectives; parliamentary decisions are discussed and deliberated on the floor in order to exchange opinions; members of the executive in most political systems are legally obliged to provide written or verbal answers to questions from legislators; and citizens express their opinions about political events on internet blogs or in public online chats. Political texts and speeches are everywhere that people express their political opinions and preferences. 

It is not until recently that social scientists have discovered the potential of analyzing political texts to test theories of political behavior. One reason is that systematically processing large quantities of textual data to retrieve information is technically challenging. Computational advances in natural language processing have greatly facilitated this task. Adaptation of such techniques in social science -- for example, Wordscore \citep{Benoit2003,Laver2003} or Wordfish \citep{Slapin2008} -- now enable researchers to systematically compare documents with one another and extract relevant information from them. Applied to party manifestos, for which most of these techniques have been developed, these methods can be used to evaluate the similarity or dissimilarity between manifestos, which can then be used to derive estimates about parties' policy preferences and their ideological distance to each other. 

One area of research that increasingly makes use of quantitative text methods are studies of legislative behavior and parliaments \citep{Giannetti2005,Laver2002,Monroe2008,Proksch2009,Yu2008,Charbonneau2009,Gallietal2009,Imbeau2009dissonance,HopkinsKing10,Quinnetal10,Grimmer10}. Only a few parliaments in the world use roll-call votes (the recording of each legislator's decision in a floor vote) that allow for the monitoring of individual members' behavior. In all other cases, contributions to debates are the only outcome that can be observed from individual members. Using such debates for social science research, however, is often limited by data availability. Although most parliaments keep written records of parliamentary debates and often make such records electronically available, they are never published in formats that facilitate social science research. A significant amount of labor is usually required to collect, clean and organize parliamentary records before they can be used for analytical purposes, often requiring technical skills that many social scientists lack. 

The purpose of this paper is to present a new database of parliamentary debates to overcome precisely this barrier. Our database contains all debates as well as questions and answers in D{\'a}il {\'E}ireann, covering almost a century of political discourse from 1919 to 2013. These debates are organized in a way that allows users to search by date, topics or speaker. More importantly, and lacking in the official records of parliamentary debates, we have identified all speakers and linked their debate contributions to the information on party affiliation and constituencies from the official members database. This enables researchers to retrieve member-specific speeches on particular topics or within a particular timeframe. Furthermore, all data can be retrieved and stored in formats that can be accessed using commonly used statistical software packages.\footnote{The database is made available on the Harvard Dataverse at \url{http://dx.doi.org/10.7910/DVN/6MZN76}.} 

In addition to documenting this database, we also present three applications in which we make use of the new data (Section \ref{sec:analysing_the_content_of_parliamentary_debates}). In the first study, we analyze budget speeches delivered by all finance ministers from 1922 to 2008 (Section \ref{sec:budget_speeches_in_historic_perspective}) and show how the policy agenda and ministers' policy preferences have changed over time (Section \ref{sec:finance_ministers_policy_positions}). In the second application we compare contributions that were made on one particular topic: the 2008 budget debate (Section \ref{sec:2008_budget_debate}). Here we demonstrate how text analytics can be used to estimate members' policy preferences on a dimension that represents pro- versus anti-government attitudes. Finally, we estimate all contributions from members of the 26th government that formed as a coalition between Fianna F{\'a}il and the Progressive Democrats in 2002. Here we estimate the policy positions of all cabinet ministers on a pro- versus anti-spending dimension and show that positions on this dimension are highly correlated with the actual spending levels of each ministerial department (Section \ref{sec:ministers_policy_position_in_the_26th_government}). 

\section{Overview of Database Content}

Parliamentary debates in D{\'a}il {\'E}ireann are collected by the Oireachtas' Debates Office and published as the Official Record. The Debates Office records and transcribes all debates and then publishes them both in printed as well as in digital form. All debates are then published on Oireachtas' website as single HTML files.\footnote{Official records are available at http://debates.oireachtas.ie/Main.aspx (last accessed on 6 August 2017). More detailed information about the Debates Office's work can be found at http://www.oireachtas.ie/viewdoc.asp?fn=/documents/Organisation/debatesoffice2.htm (last accessed on 6 August 2017).} At the time of writing, the official debates website contains 549,292 HTML files. The content of all these HTML files forms the data source for our database. It is obviously impossible to hand-code that much information. We therefore wrote a computer script that automated the processing of all files.\footnote{The computer script consists of multiple syntax files that were written in Python.} This script is able to find all debate contributions and the names of all speakers in each file. In addition, it retrieves the date as well as the topic of each debate. 

As already explained above, the official online version of the Official Records does not provide information about speakers besides their name. Each speaker's name is ``hard coded'' into the HTML files and not linked to the information in the official members database. In addition, speaker names are not coded consistently, hence making it difficult to collect speeches from a particular deputy.\footnote{More recent parliamentary debates are made available in a dynamic framework on the Oireachtas website. This new interface allows for the retrieval of speaker specific information and to retrieve speeches from a single member. However, this only applies to recent debates since 2007, and not historical debates.} Our goal was to identify every single speaker name that appears in the Official Record and integrate parliamentary speeches with information about deputies' party affiliation, constituency, age and profession from the official members database into a single database. We therefore used an automated record-linkage procedure to identify every single speaker.\footnote{Record-linkage is a common technique that is used to link entries from two databases that share the same content but differ in how entries are coded. The basic idea of this procedure is to compare every entry from one database (in our case, the complete list of all speaker names) with every entry from the second database (in our case, the official members database), using some pre-defined algorithms to determine which two entries are most similar to one another. Different record-linkage algorithms have been developed and, after comparing several algorithms, we found the ``longest common sub-string'' procedure to work particularly well with our data. (See \cite{Christen2006} for an overview and comparison of different record-linkage procedures.) The computer code we applied comes from \emph{Febrl}, a Python environment that was developed by the ANU Data Mining Group at the Australian National University (https://sourceforge.net/projects/febrl/).}

The final database contains all debates and written answers from the first meeting of the D{\'a}il on 21 January 1919 through to 28 March 2013, covering every D{\'a}il session that has met during this period. In total, the database contains 4,443,713 individual contributions by 1,178 TDs. The data is organized in a way that facilitates analysis for substantive questions of interest to social scientists. Every row in the data set is one contribution with columns containing information on the following variables:
 
\begin{itemize}
\item the name and surname of the speaker,
\item the unique ID the speaker,
\item the speaker's party affiliation and constituency,
\item the title of the D{\'a}il debate as recorded in the official records,
\item the date of the debate.
\end{itemize}

\section{Analyzing the Content of Parliamentary Debates}
\label{sec:analysing_the_content_of_parliamentary_debates}
In the previous section, we have explained the structure of the database. In the following three sections we demonstrate how the data can be used for social science research. We do this by demonstrating three different applications. In the first application, we analyze the budget speeches of all finance ministers from 1922 to 2008. Budget speeches are delivered by Finance Ministers once a year, with the exception of emergency budgets. Analyzing this data, we show how policy agendas and ministers' fiscal preferences have changed over time. In the second application, we construct a data set that resembles a cross-sectional analysis as we retrieve all speeches from one particular year and on one particular topic from our database: the 2008 budget debate. This data structure enables us to estimate the policy positions of all speakers who contributed to the budget debate and to compare how similar or dissimilar their preferences were. We find that policy positions are clustered into two groups: the government and the opposition; but we also find considerable variation within each group. Finally, we take all contributions made during the term of one government and use the data to estimate the policy positions of all cabinet members on a dimension representing pro- versus anti-spending. We demonstrate the validity of estimated policy positions by comparing them against actual spending levels of each cabinet ministers' department and show that the two measures are almost perfectly correlated with each other.

\subsection{The Content of Budget Speeches in Historical Perspective}
\label{sec:budget_speeches_in_historic_perspective}
The quantitative analysis of text is primarily based on the proposition that preference profiles of speakers can be constructed from their word frequencies \citep{baayen2001,Bybee2001}. This makes word frequencies the most important data input to almost all existing methods of text analysis. Word frequencies can be easily visualized as \emph{word clouds}. These word clouds show the most frequently used words in a text with font size being proportional to frequency of appearance. Despite their simplicity, word clouds can be used as a first descriptive view of the data. Here we look at word clouds for the speeches made by Irish Ministers for Finance. We have extracted the budget speeches of all finance ministers from our database, the first being Cosgrave's speech in April 1923, and the latest being Lenihan's speech in October 2008. In total, there are 90 speeches given by 23 different finance ministers for whom we have generated word clouds as shown in Figure \ref{fig:word_clouds}. 

One way to look at Figure \ref{fig:word_clouds} is to consider that each individual word cloud panel presents a snapshot into the preference profiles of individual ministers. With taxation being the key instrument of fiscal policy it is unsurprising that the word ``tax'' is on average the most frequently used word across all Ministers for Finance. We can also discern that frequency of references to ``government'' has been uneven over time with relatively high usage in the 1960s to 1980s and then subsequent decline (apart from Quinn's tenure) until the later speeches of Cowen and particularly Lenihan. 

What is more clearly evident is the change in the number of unique words used by different ministers. This reflects the fact that some budget speeches were very short, while others were long and covered many distinct topics. The easiest example is to compare speeches by two consecutive ministers: Cowen and Lenihan. Word clouds reflect the sheer multitude of problems facing the country that needed to be addressed by Lenihan compared to the relatively ``quieter'' (on average) three budgets delivered by Cowen.

\begin{figure}
\begin{center}
\begin{minipage}{\textwidth}
\fbox{\includegraphics[width=.49\textwidth]{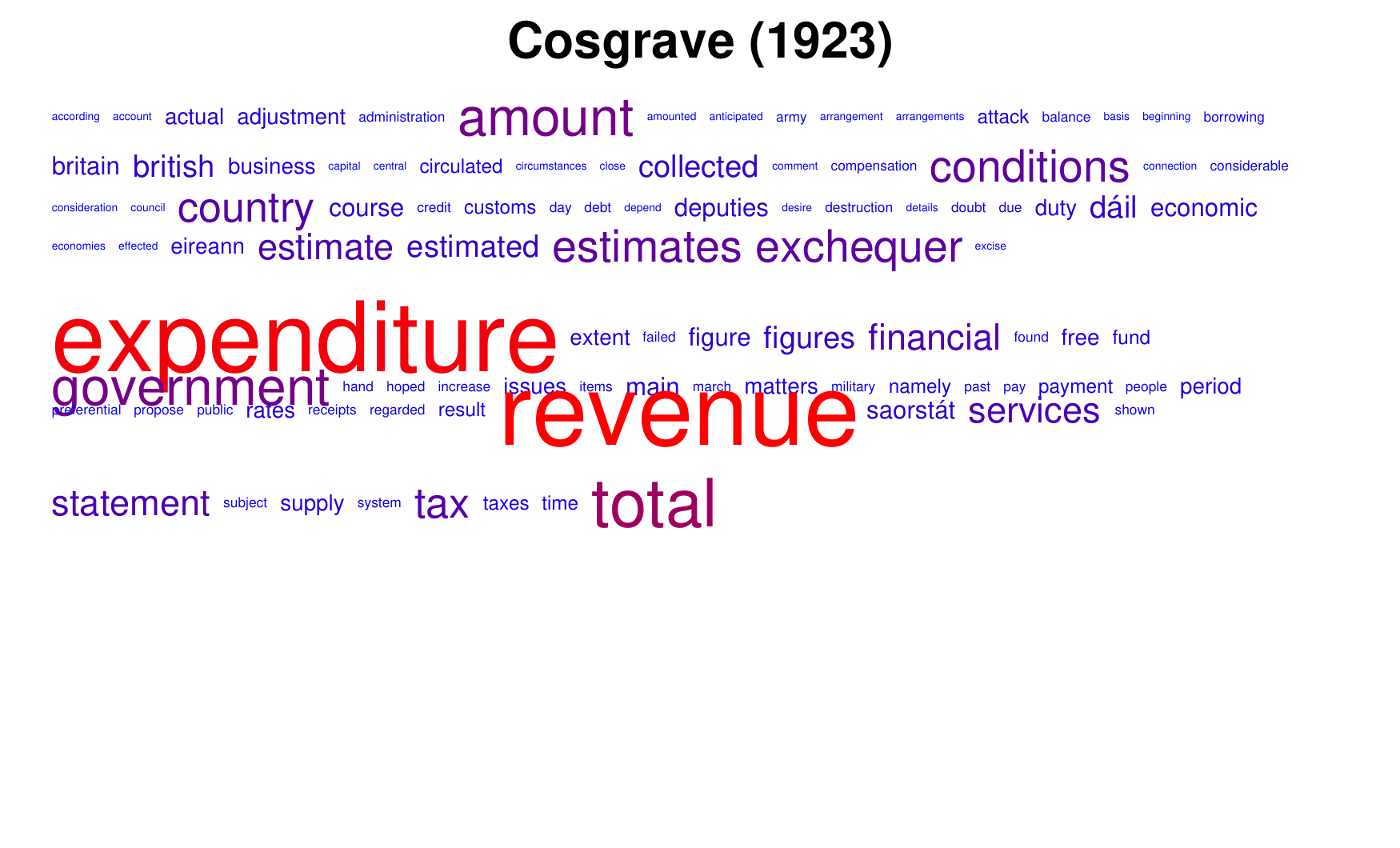}} 
\fbox{\includegraphics[width=.49\textwidth]{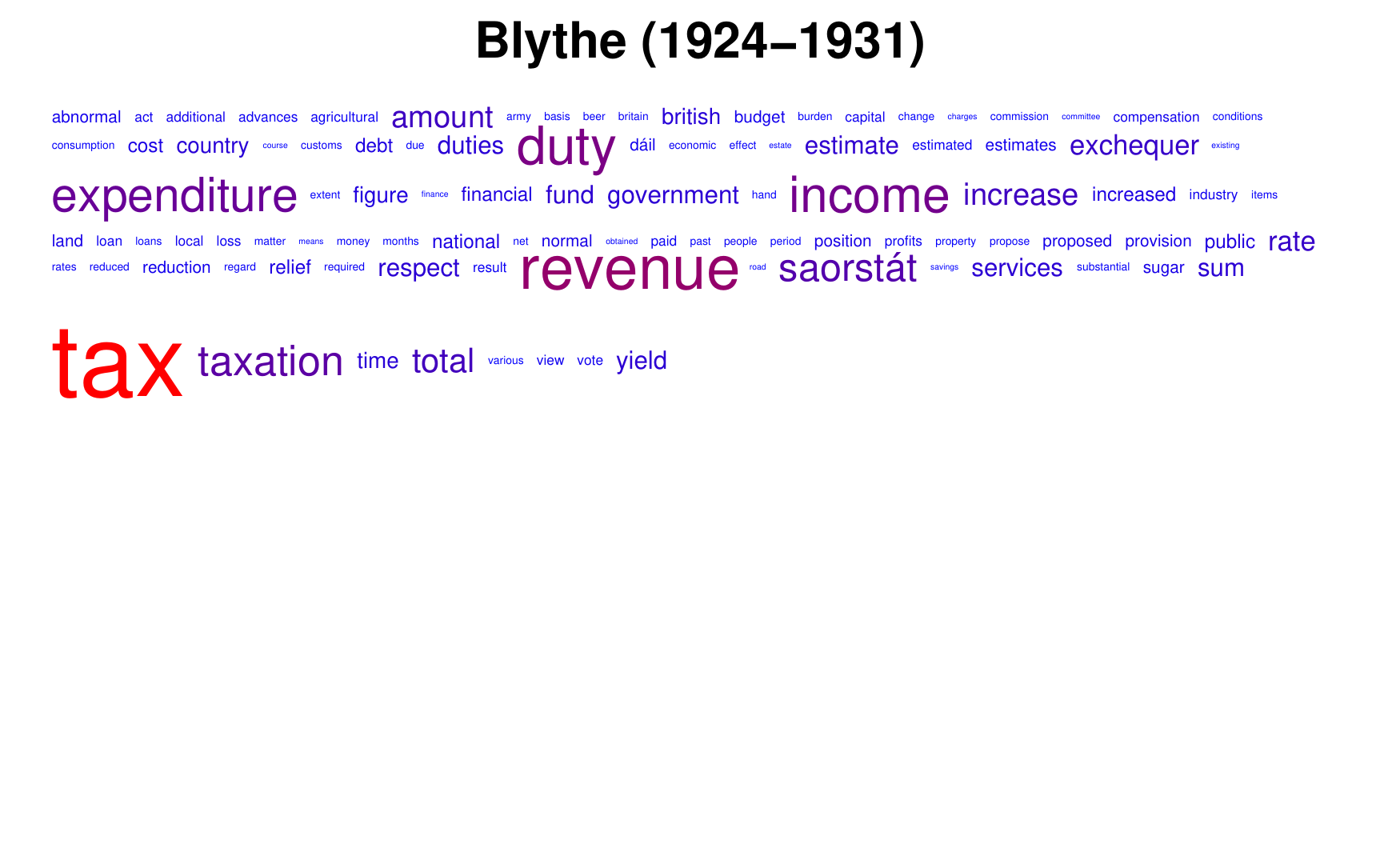}} 
\fbox{\includegraphics[width=.49\textwidth]{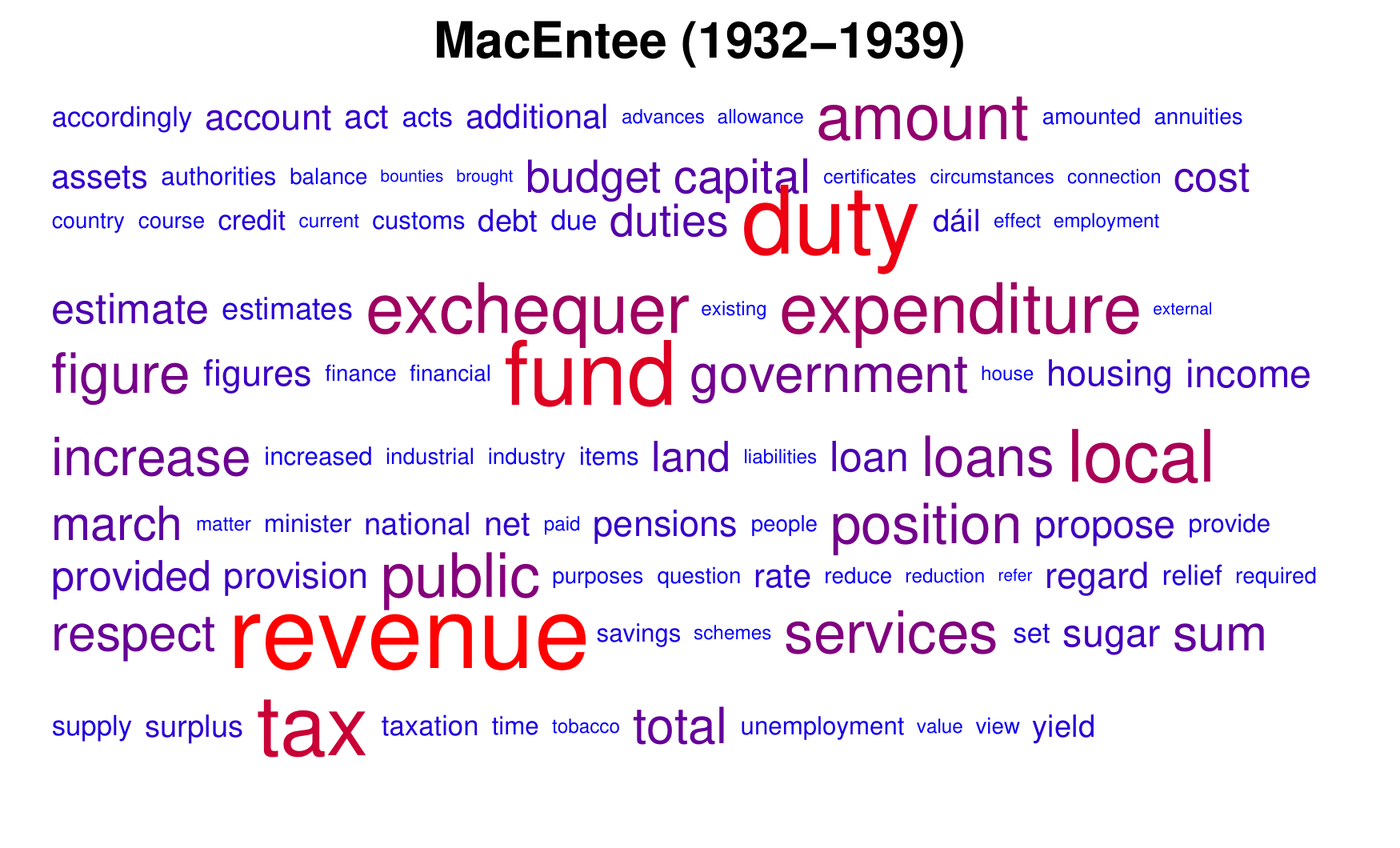}} 
\fbox{\includegraphics[width=.49\textwidth]{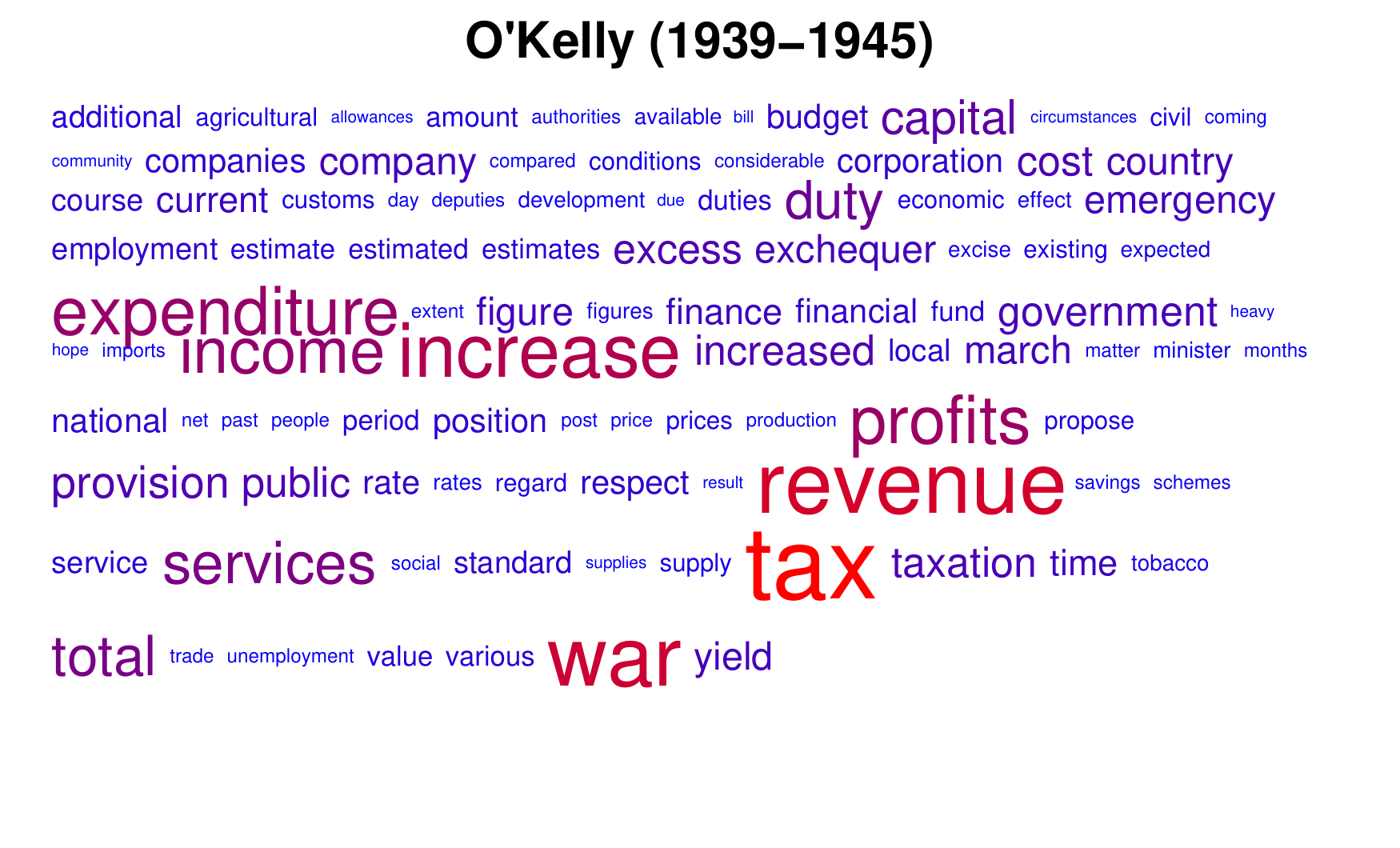}} 
\fbox{\includegraphics[width=.49\textwidth]{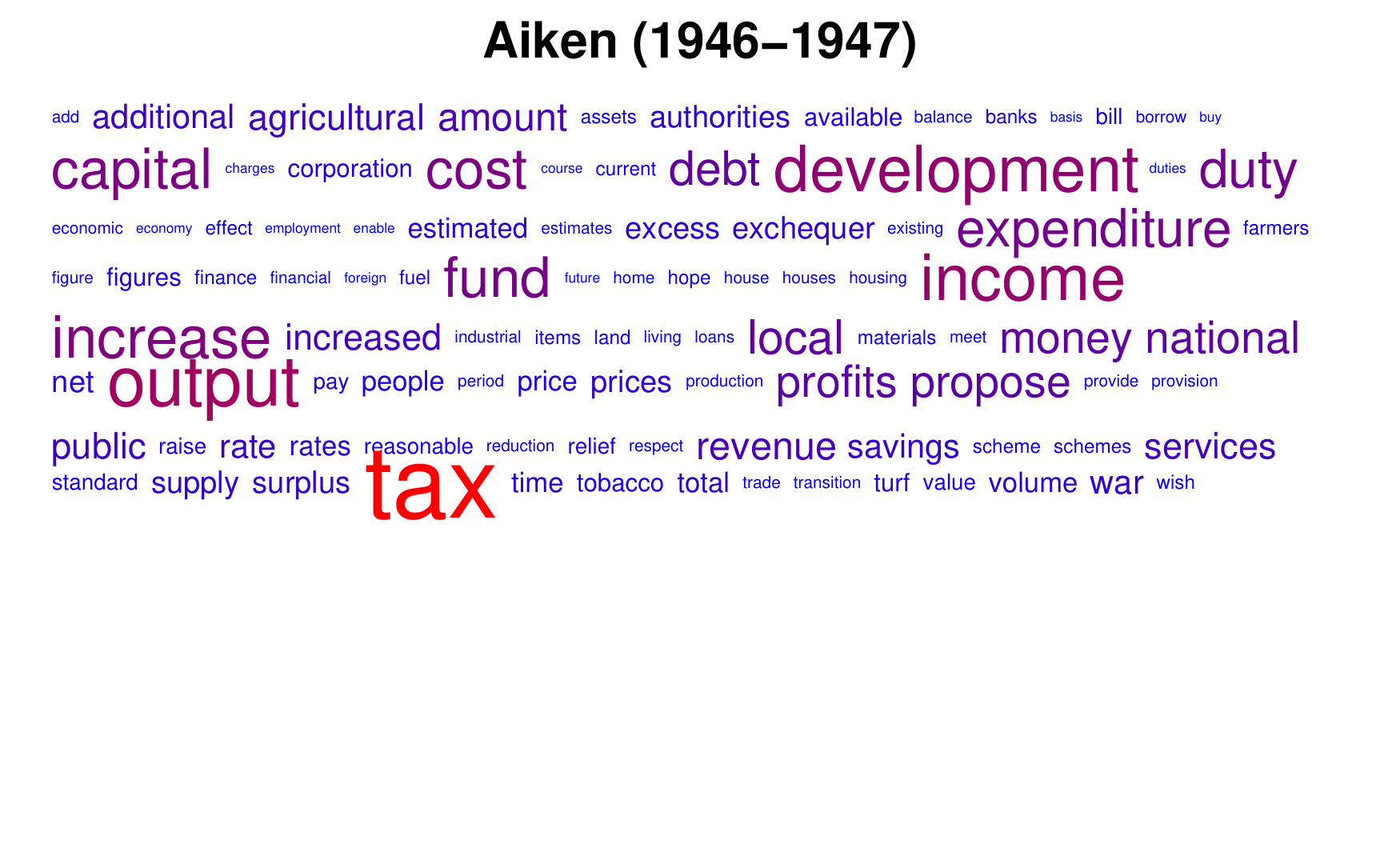}} 
\fbox{\includegraphics[width=.49\textwidth]{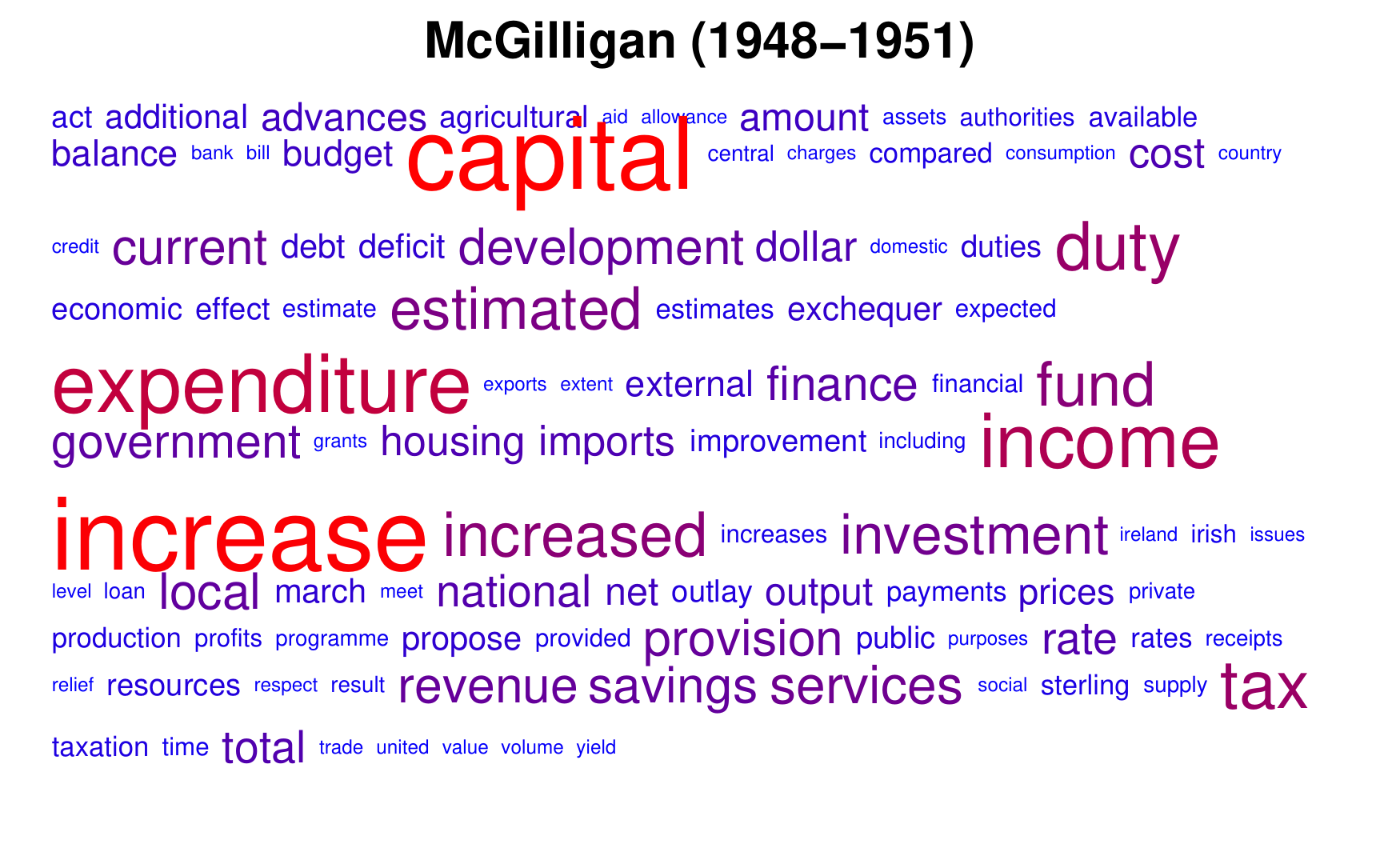}} 
\fbox{\includegraphics[width=.49\textwidth]{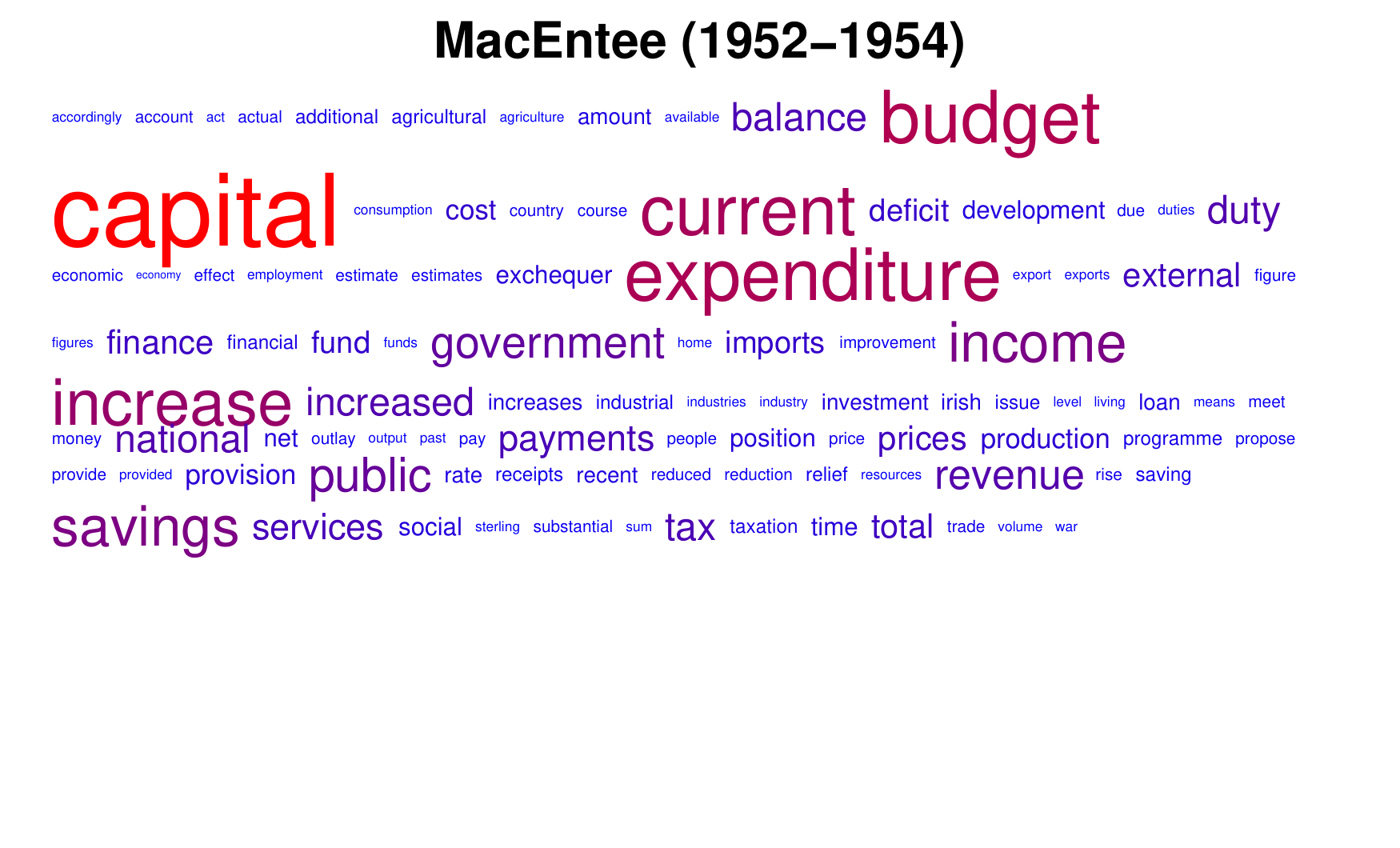}} 
\fbox{\includegraphics[width=.49\textwidth]{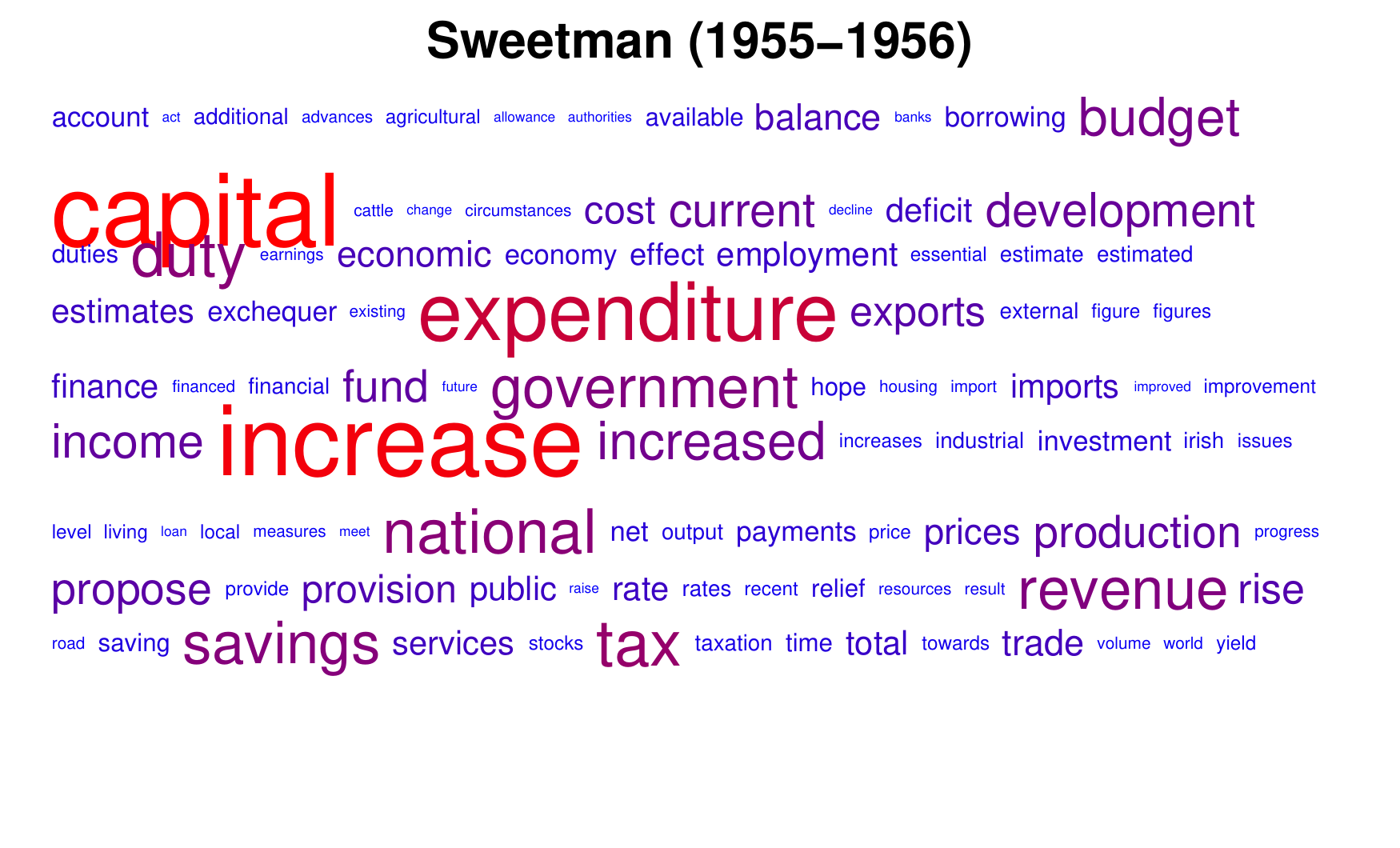}} 
\end{minipage}
\end{center}
\caption{\emph{Word clouds of all Finance Ministers' budget speeches, 1922--2008} \label{fig:word_clouds}}
\end{figure}

\begin{figure}
\begin{center}
\begin{minipage}{\textwidth}
\fbox{\includegraphics[width=.49\textwidth]{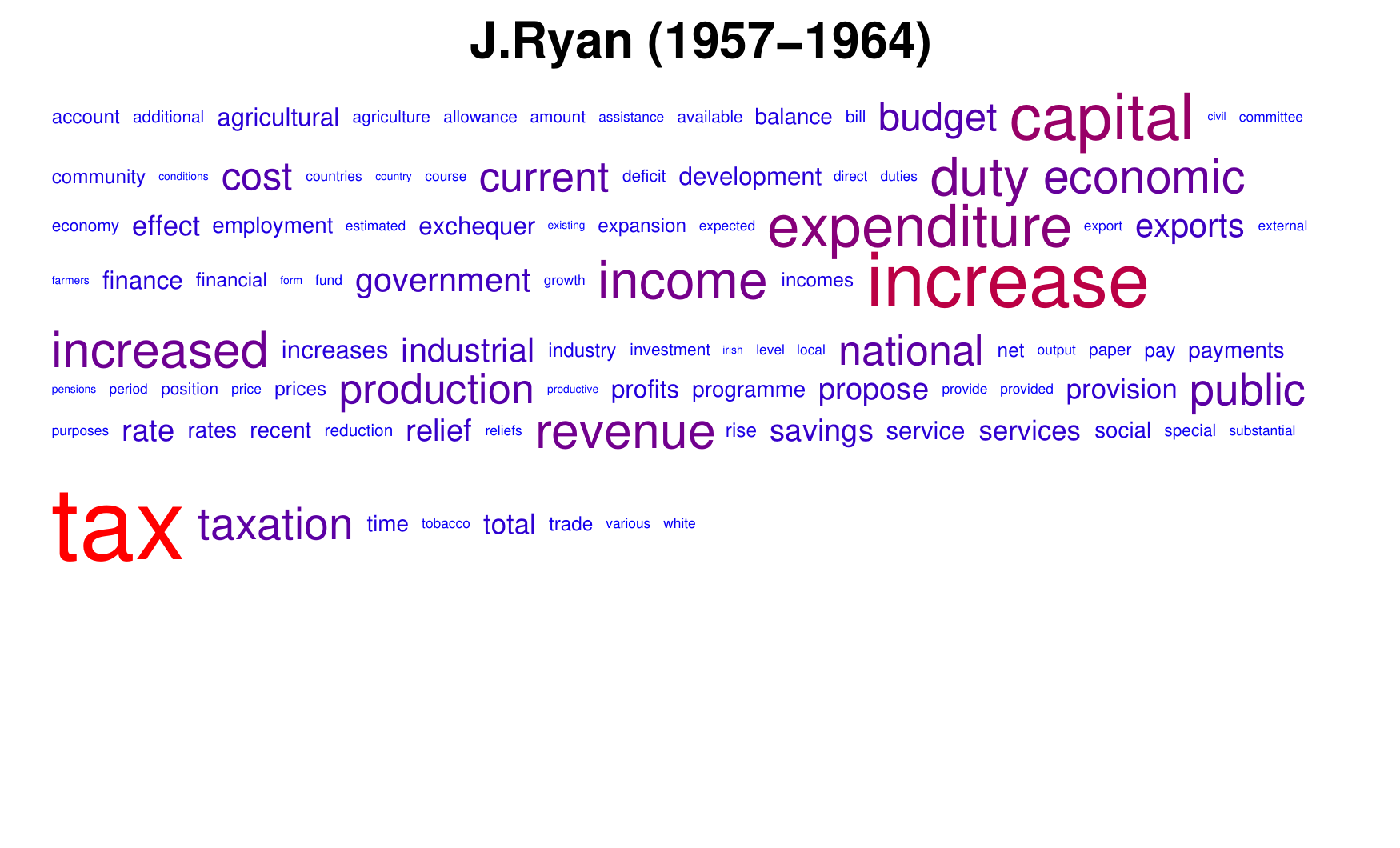}}
\fbox{\includegraphics[width=.49\textwidth]{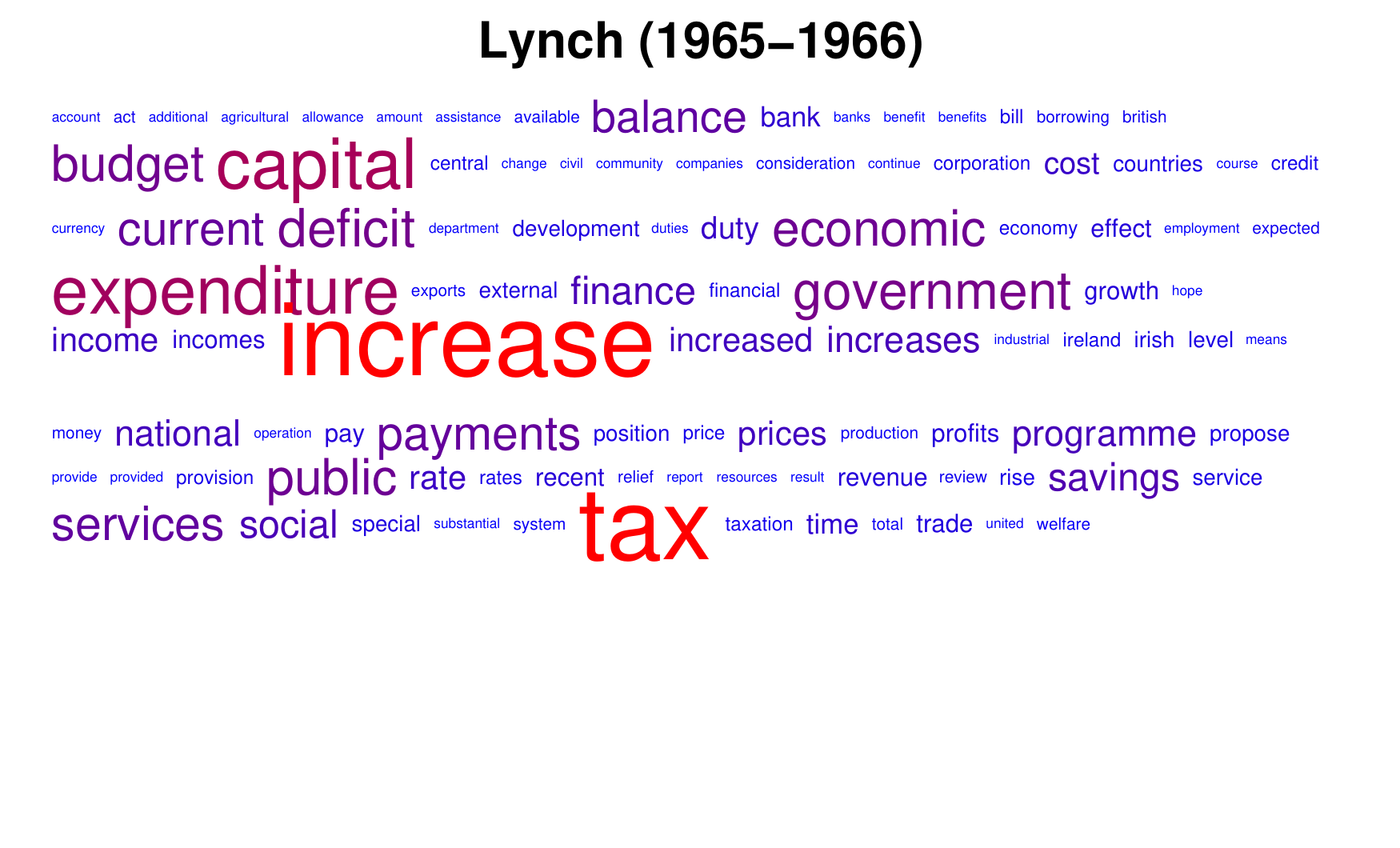}} 
\fbox{\includegraphics[width=.49\textwidth]{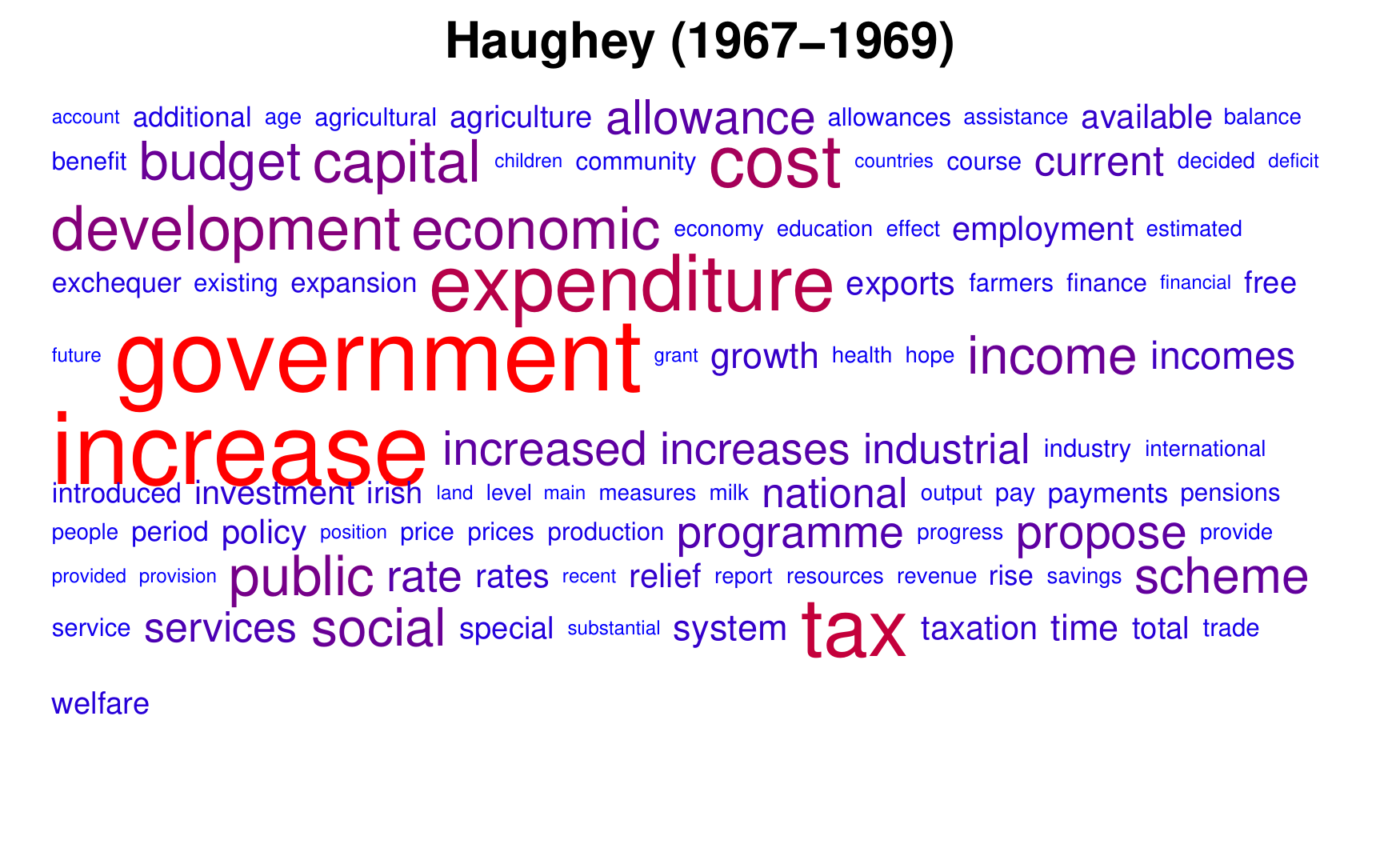}} 
\fbox{\includegraphics[width=.49\textwidth]{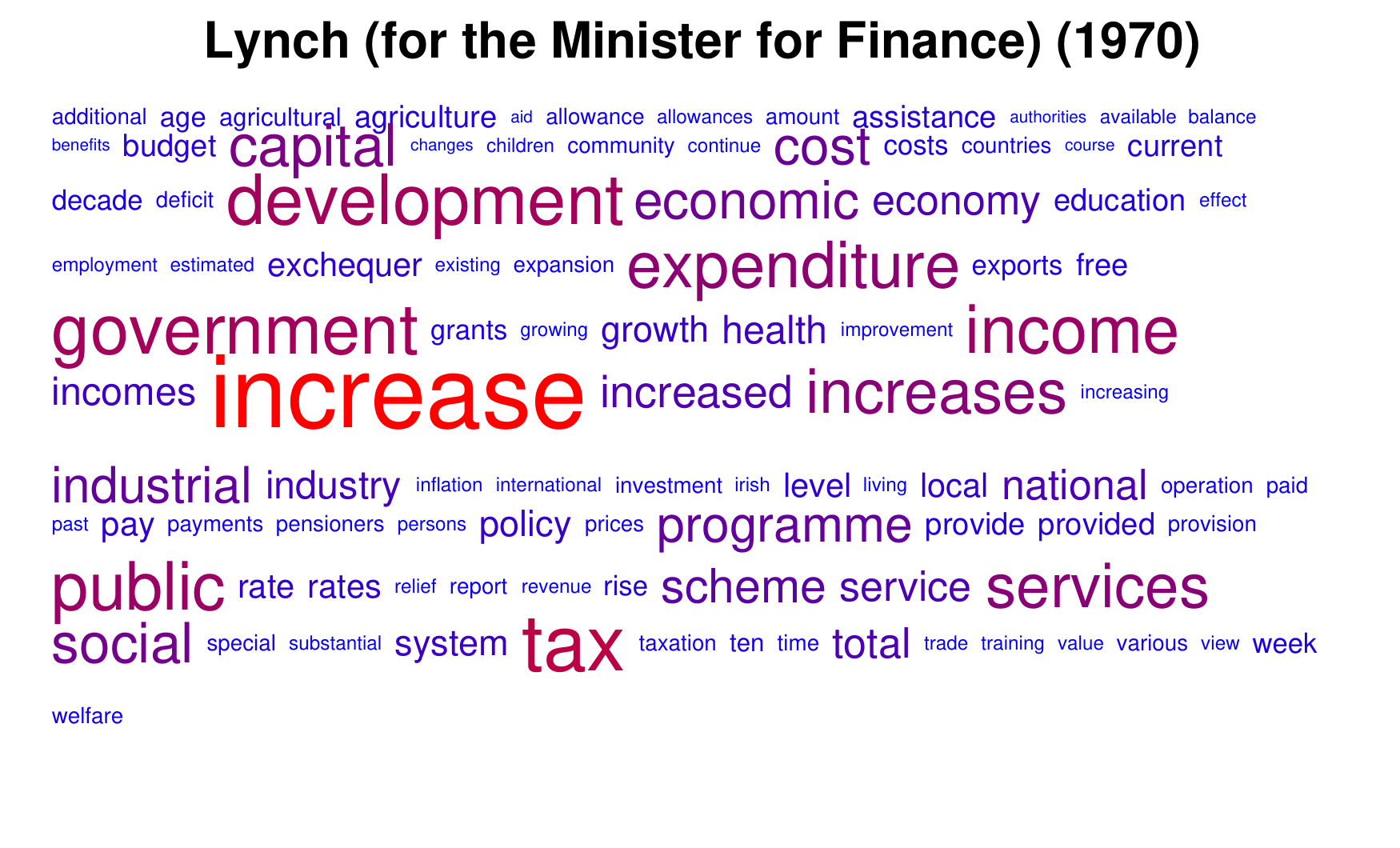}} 
\fbox{\includegraphics[width=.49\textwidth]{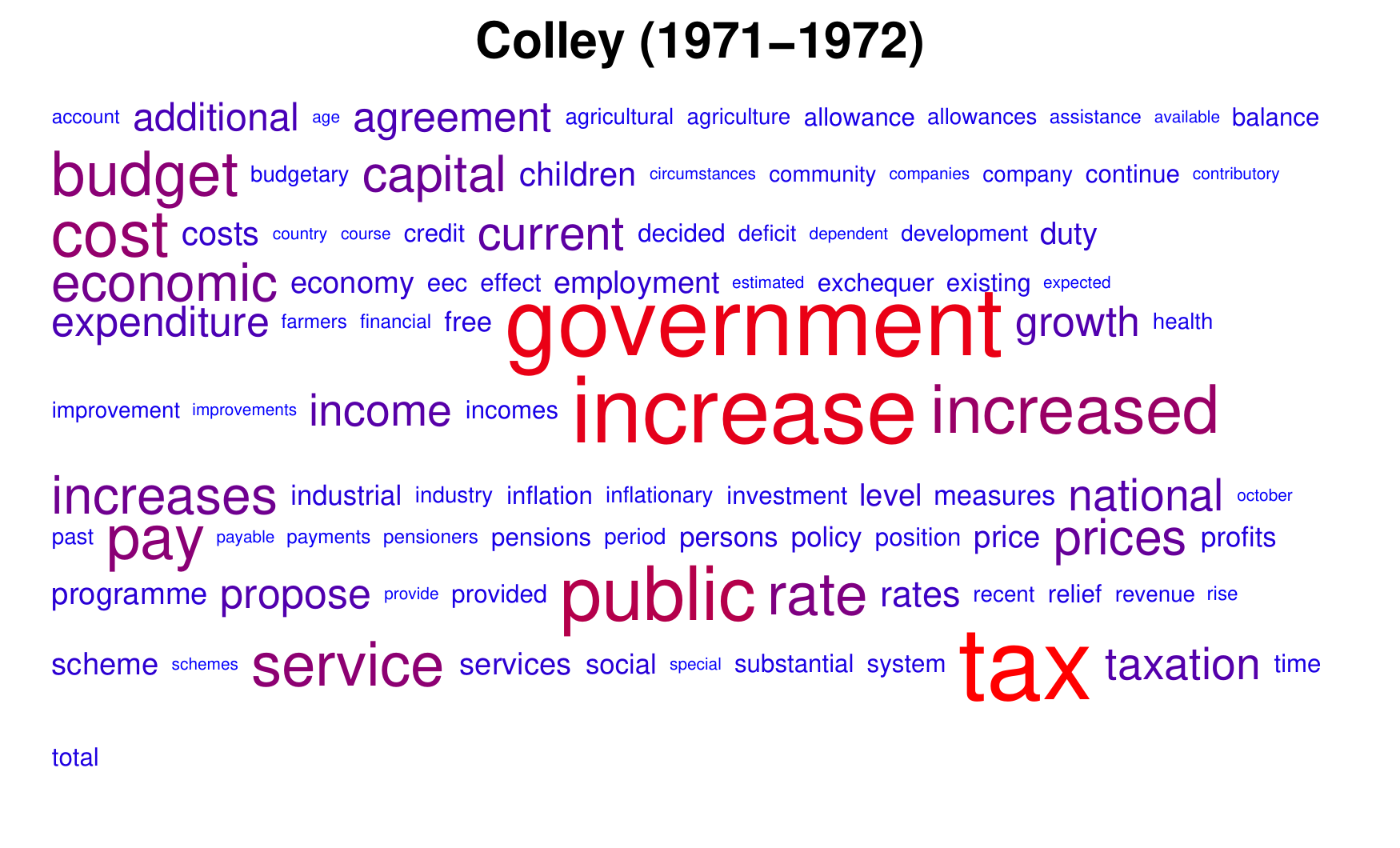}} 
\fbox{\includegraphics[width=.49\textwidth]{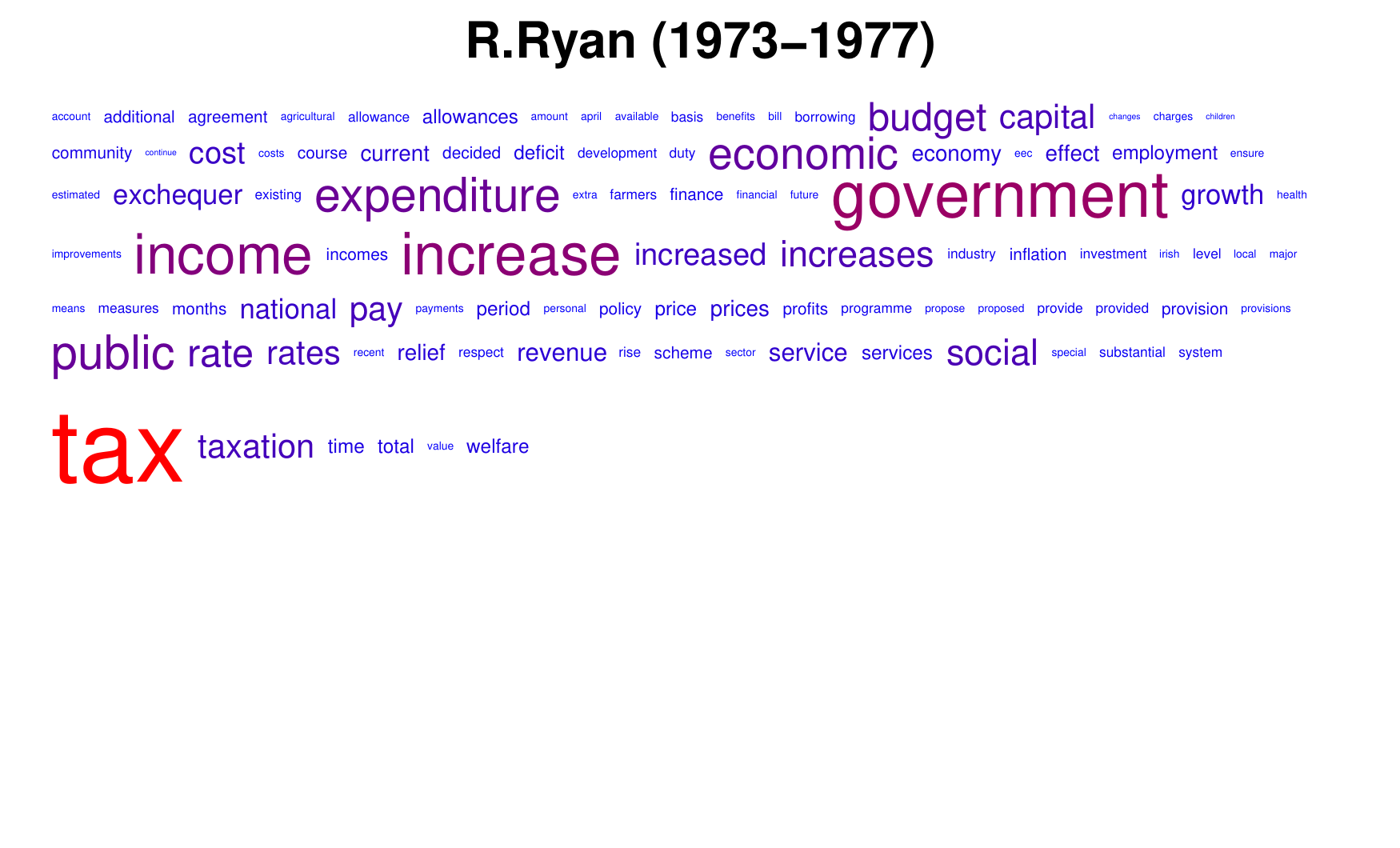}} 
\fbox{\includegraphics[width=.49\textwidth]{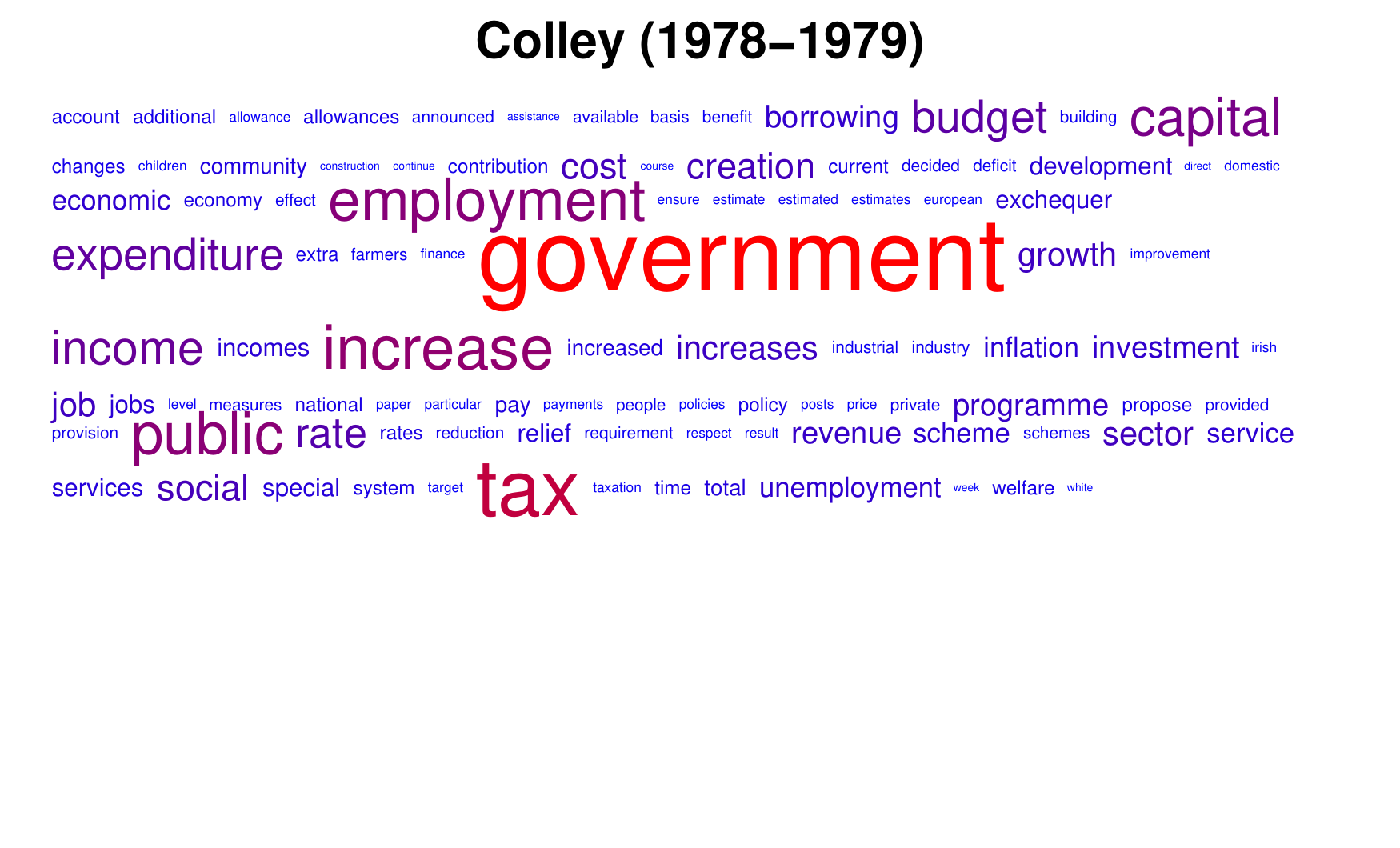}} 
\fbox{\includegraphics[width=.49\textwidth]{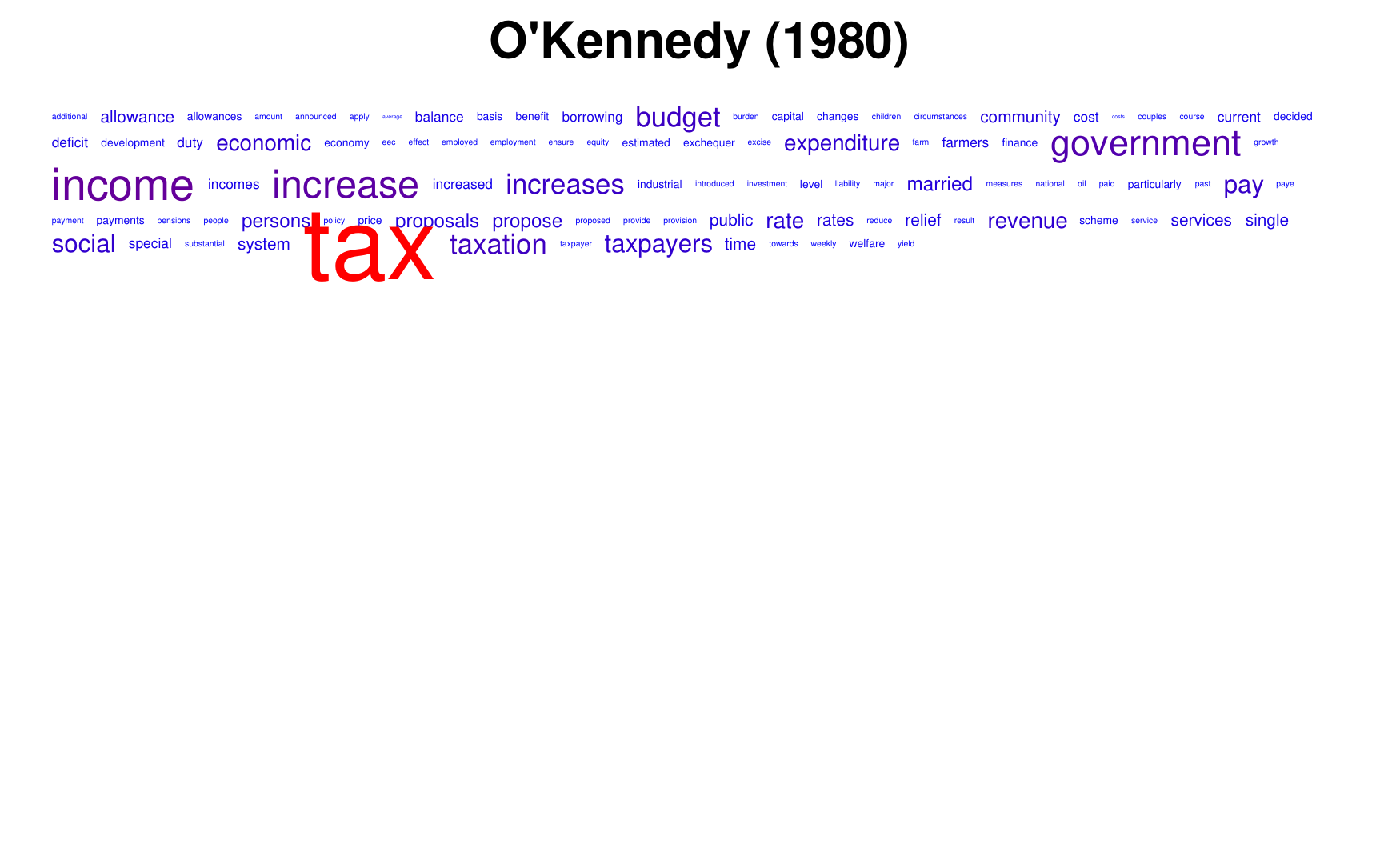}} 
\end{minipage}
\end{center}
\addtocounter{figure}{-1} 
\caption{\emph{Word clouds of all budget speeches made by Ministers for Finance, 1922--2008 (cont'd)}.}
\end{figure}

\begin{figure}
\begin{center}
\begin{minipage}{\textwidth}
\fbox{\includegraphics[width=.49\textwidth]{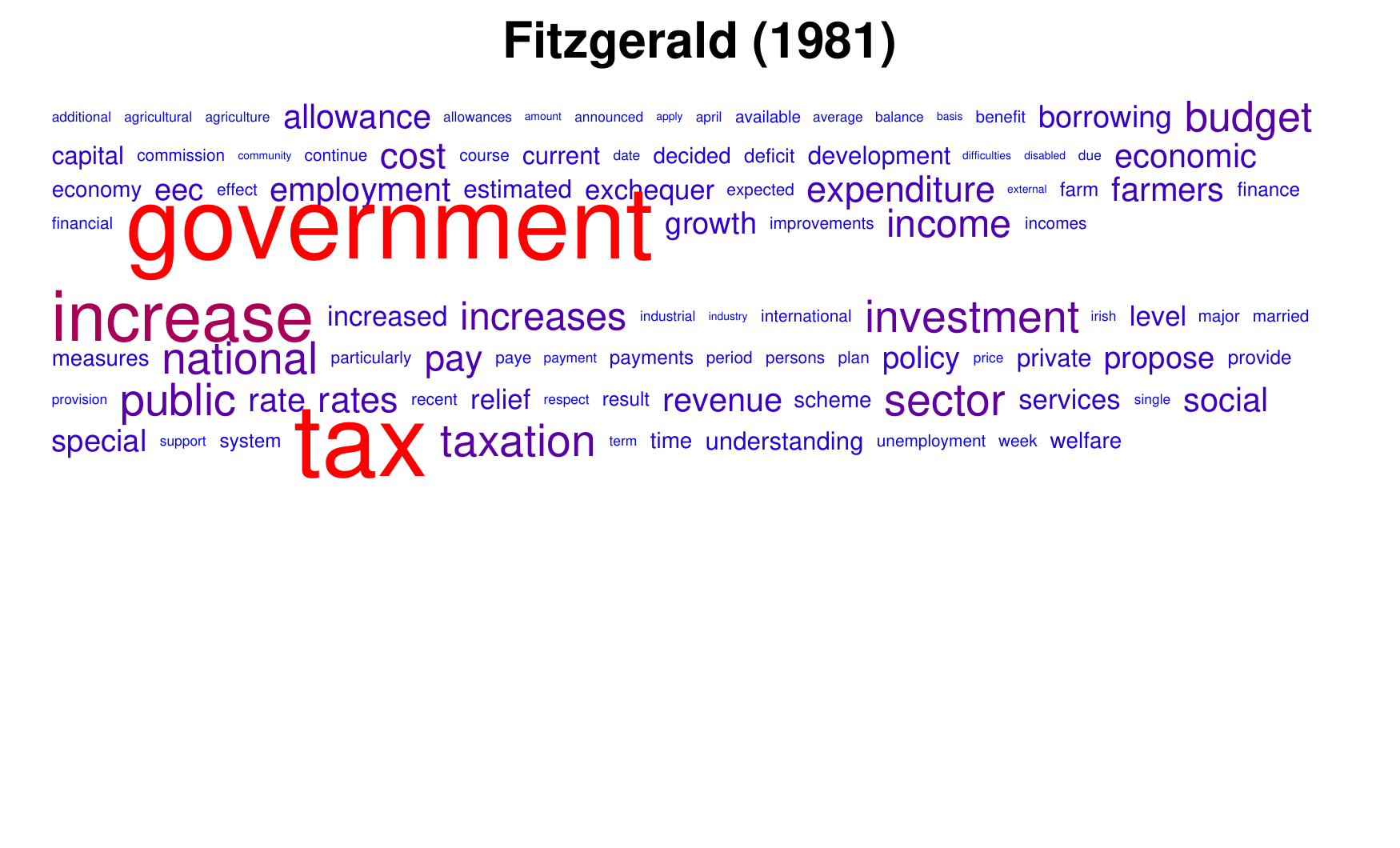}} 
\fbox{\includegraphics[width=.49\textwidth]{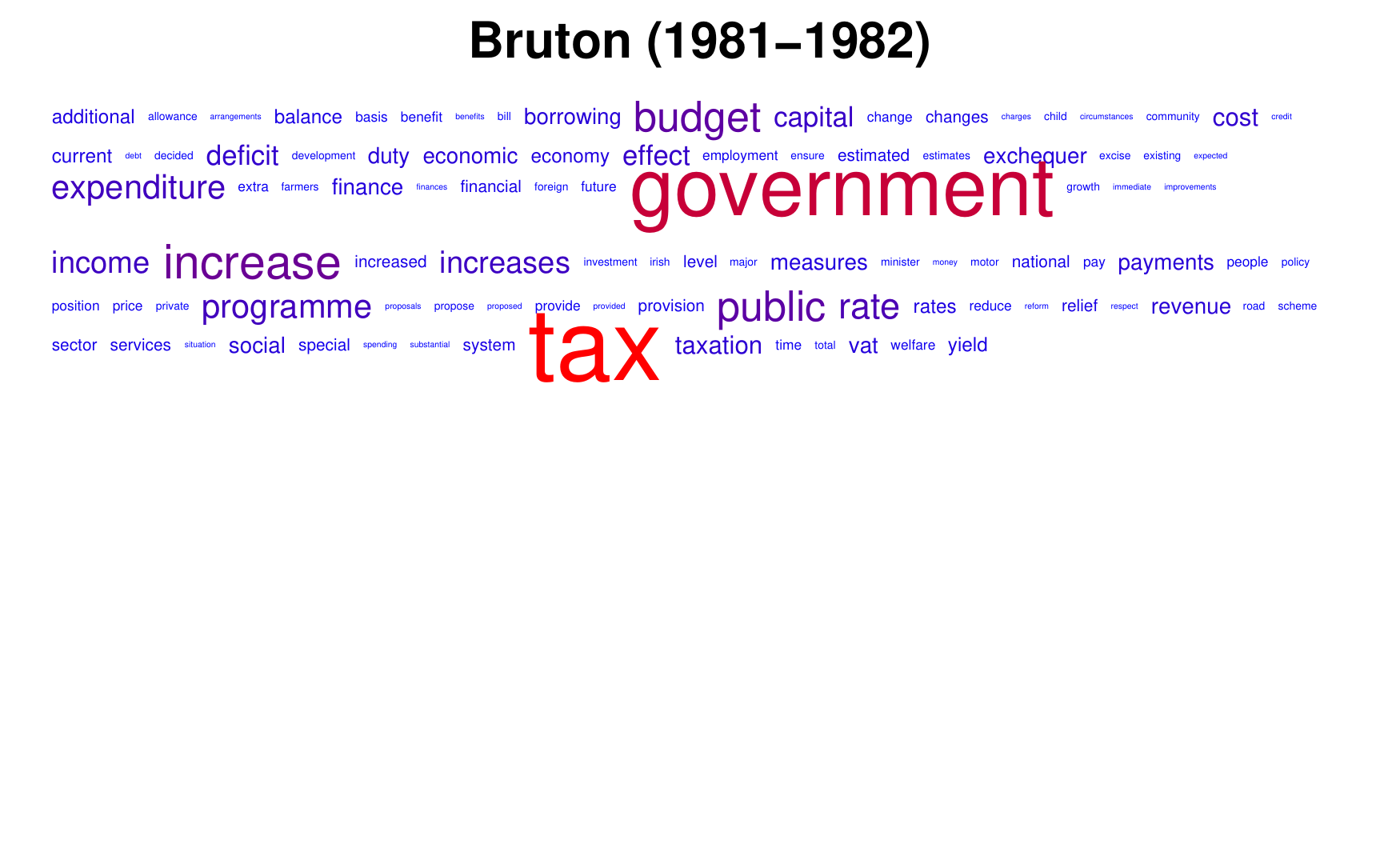}} 
\fbox{\includegraphics[width=.49\textwidth]{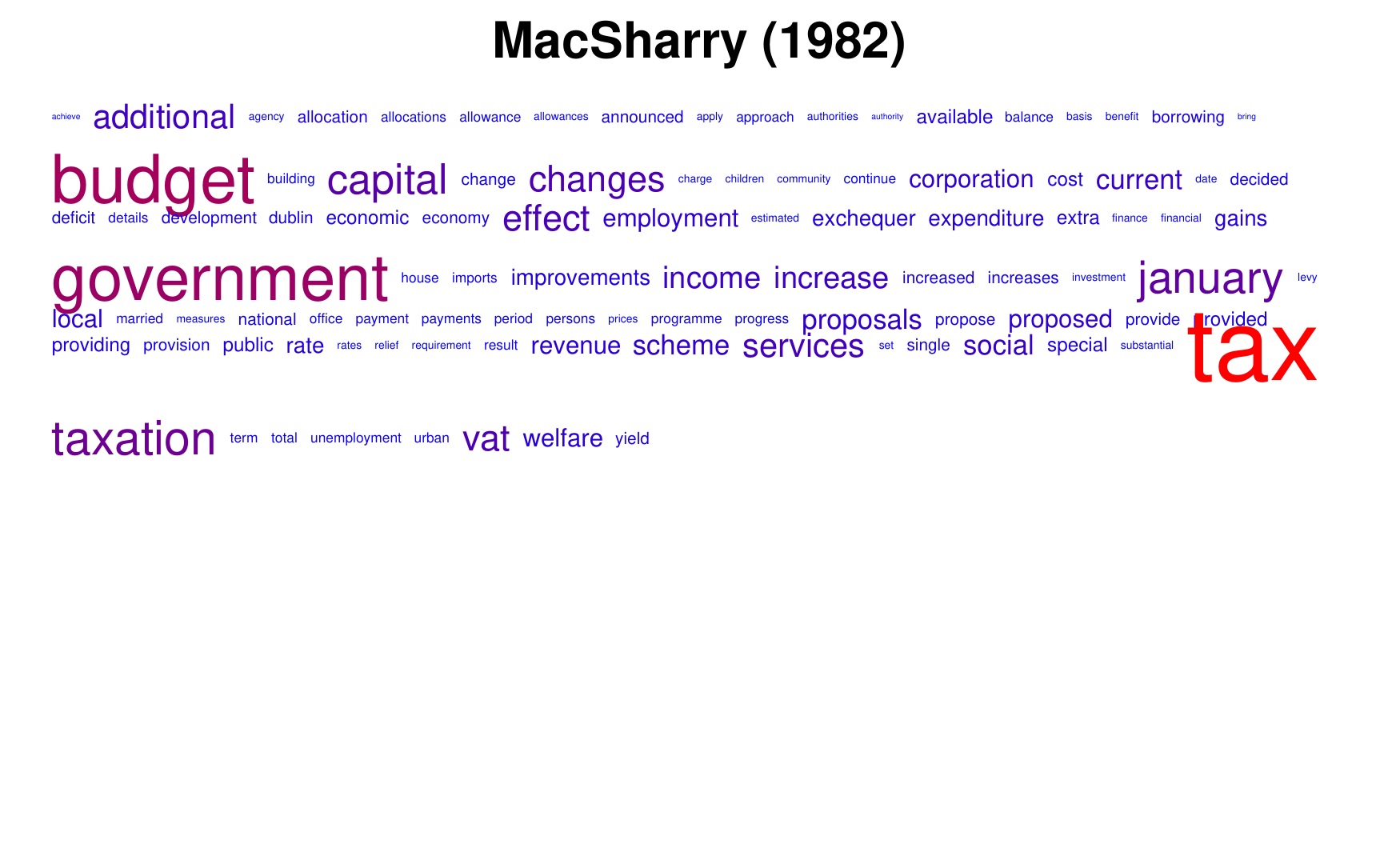}} 
\fbox{\includegraphics[width=.49\textwidth]{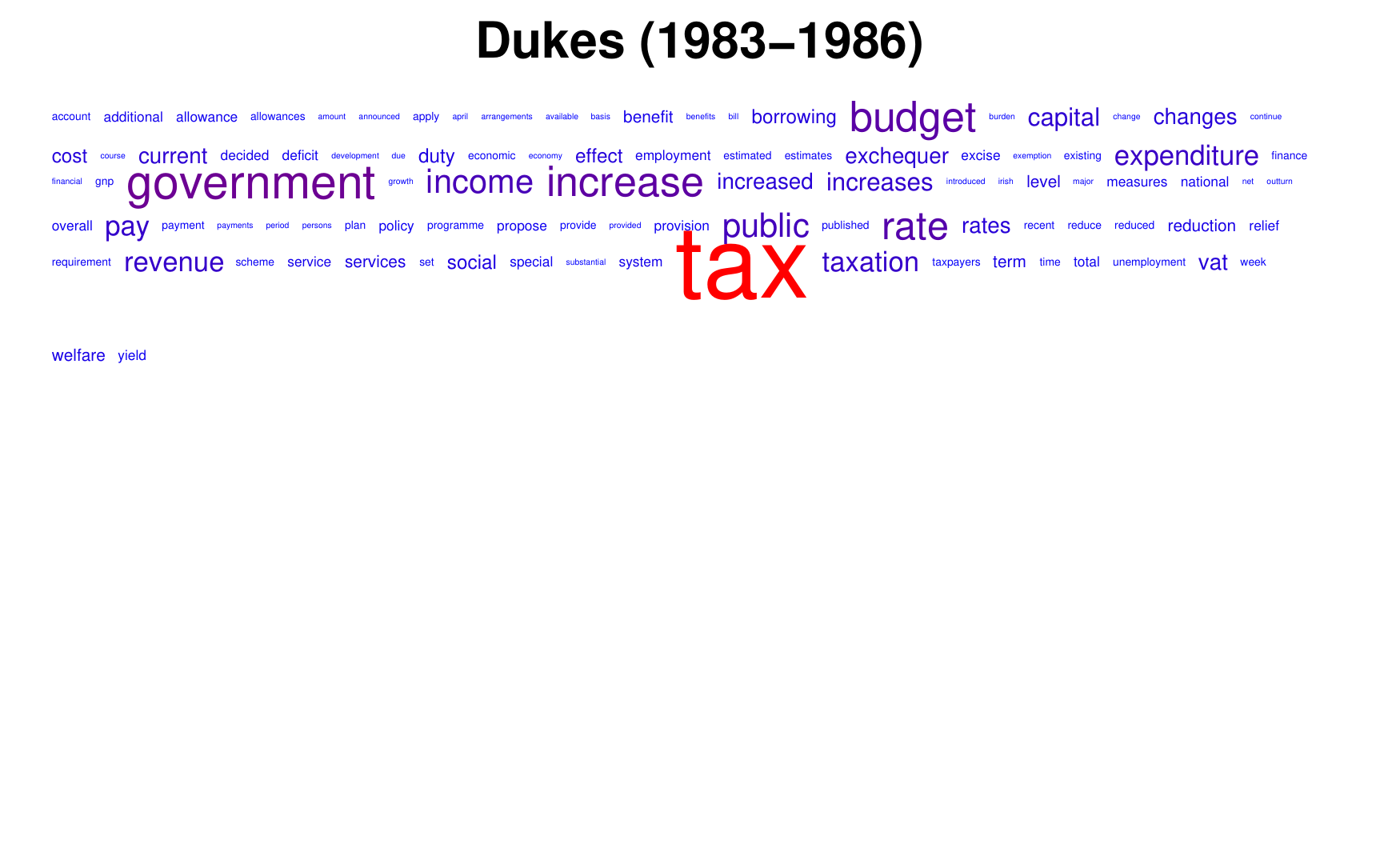}} 
\fbox{\includegraphics[width=.49\textwidth]{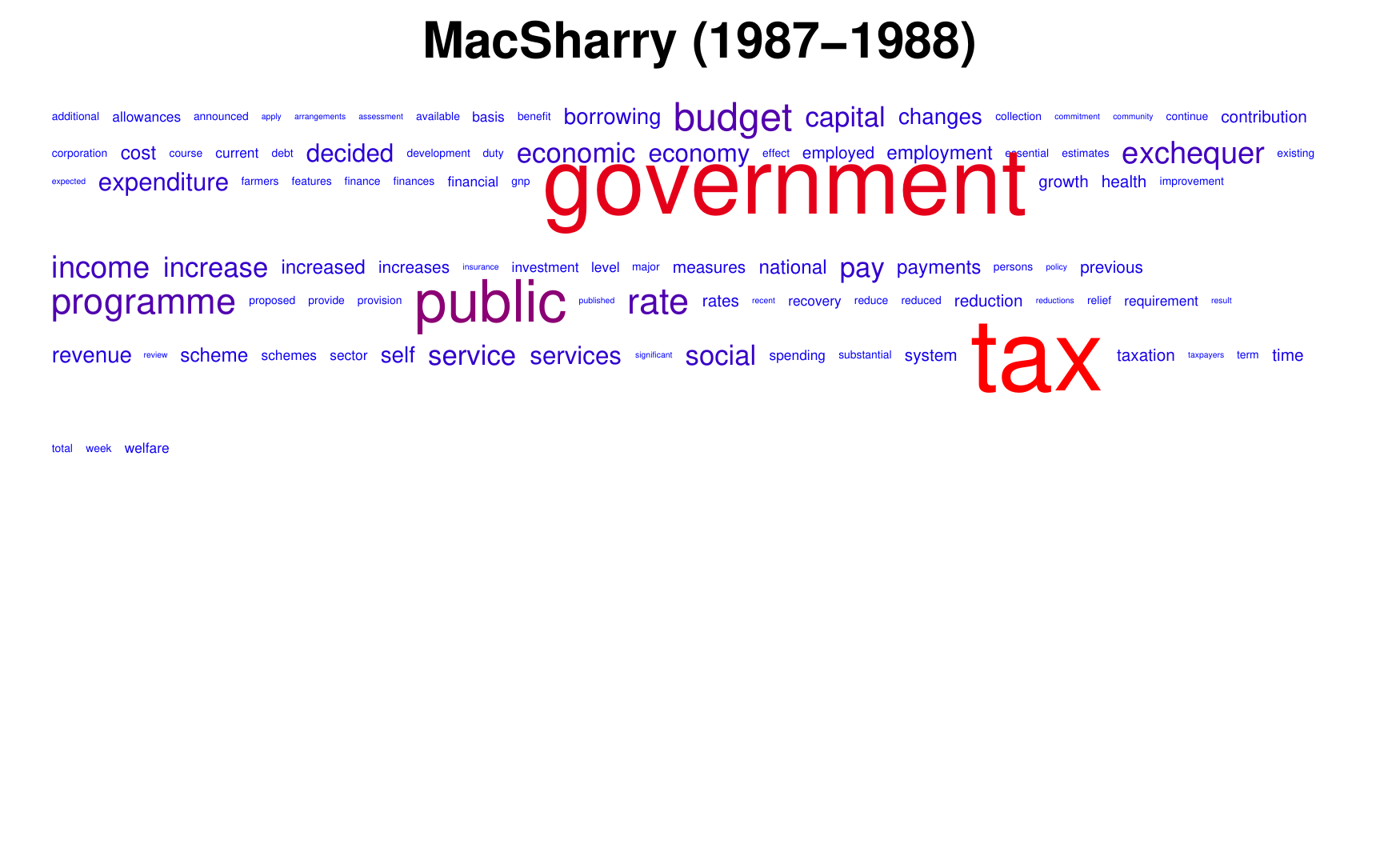}} 
\fbox{\includegraphics[width=.49\textwidth]{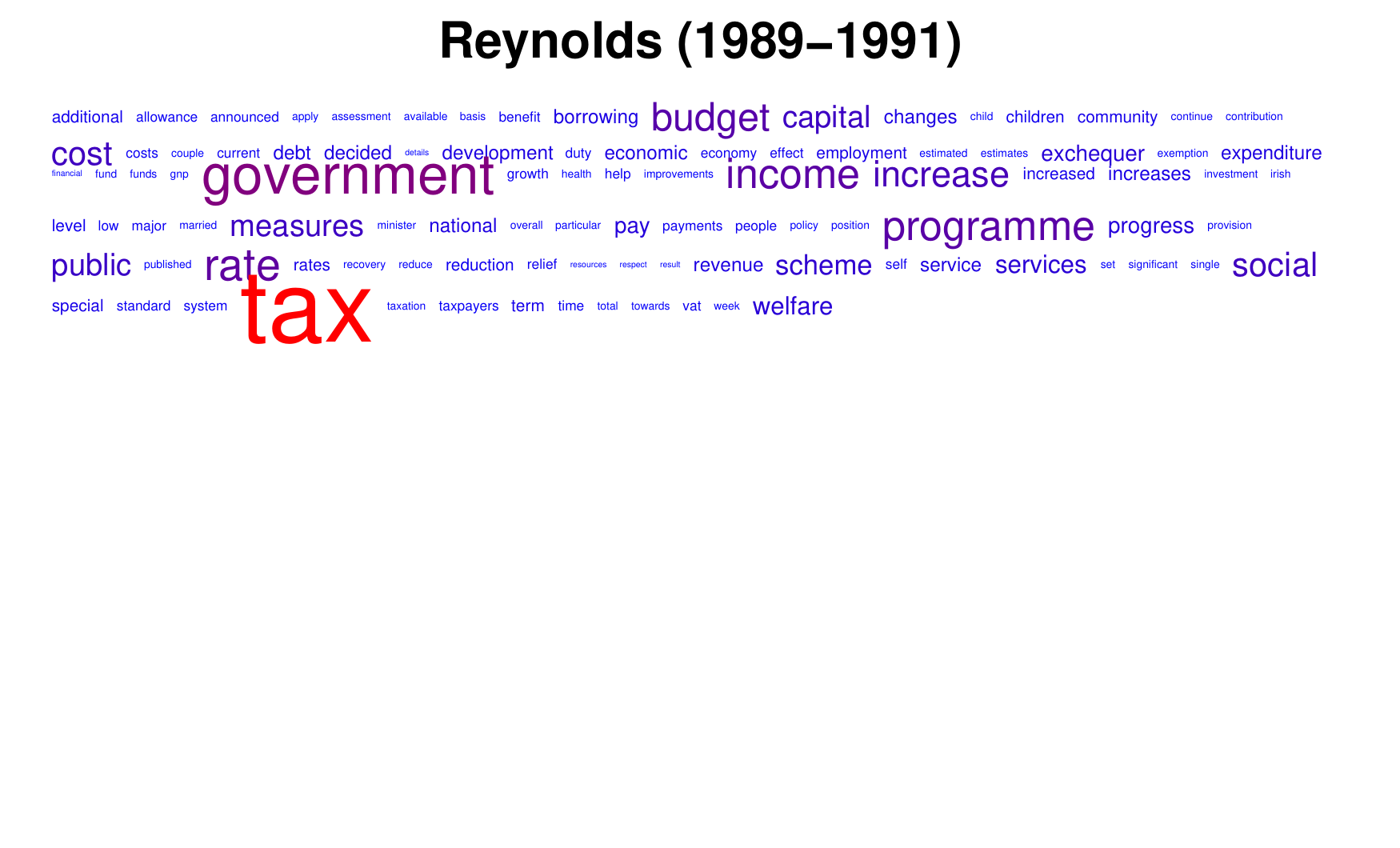}} 
\fbox{\includegraphics[width=.49\textwidth]{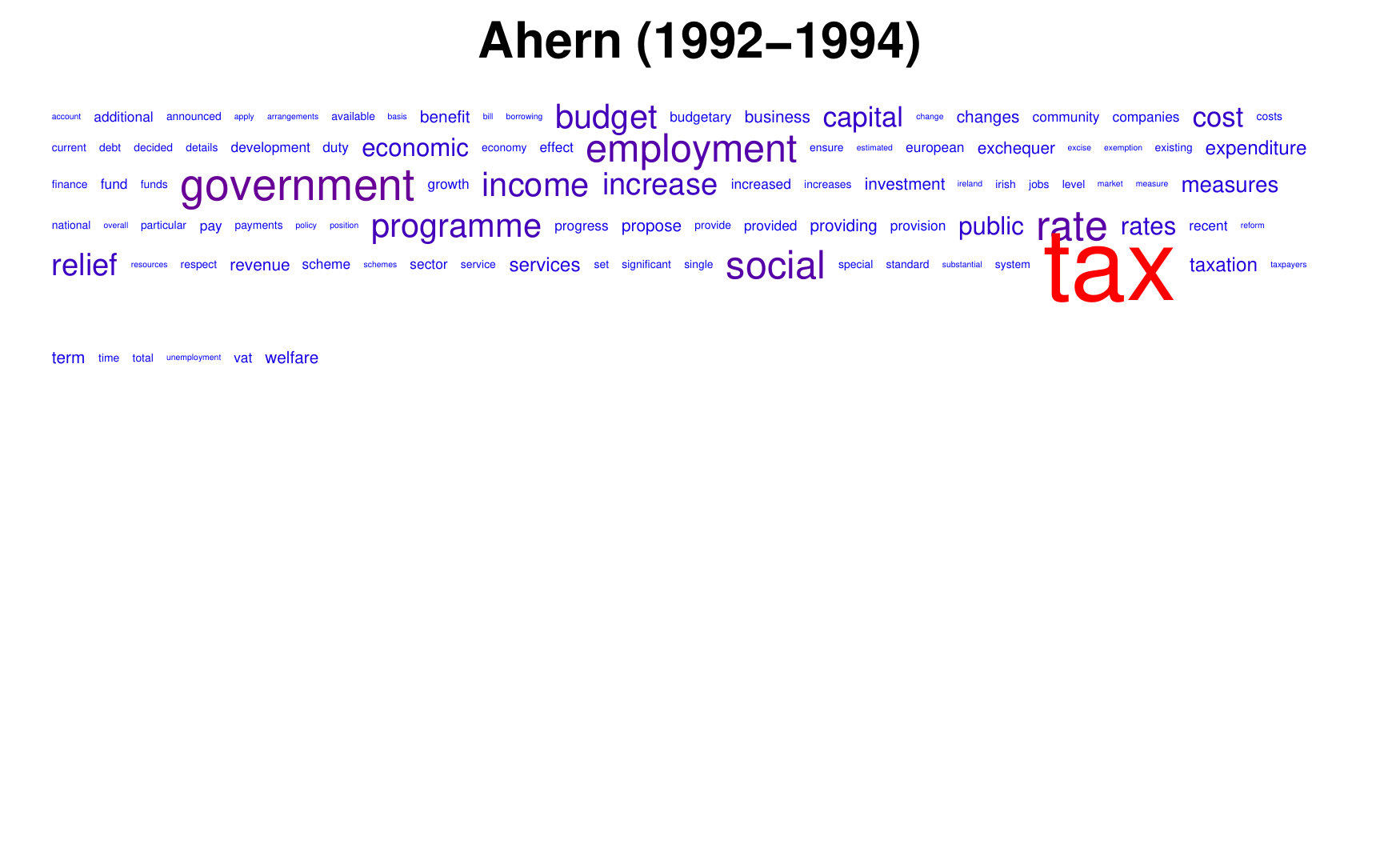}} 
\fbox{\includegraphics[width=.49\textwidth]{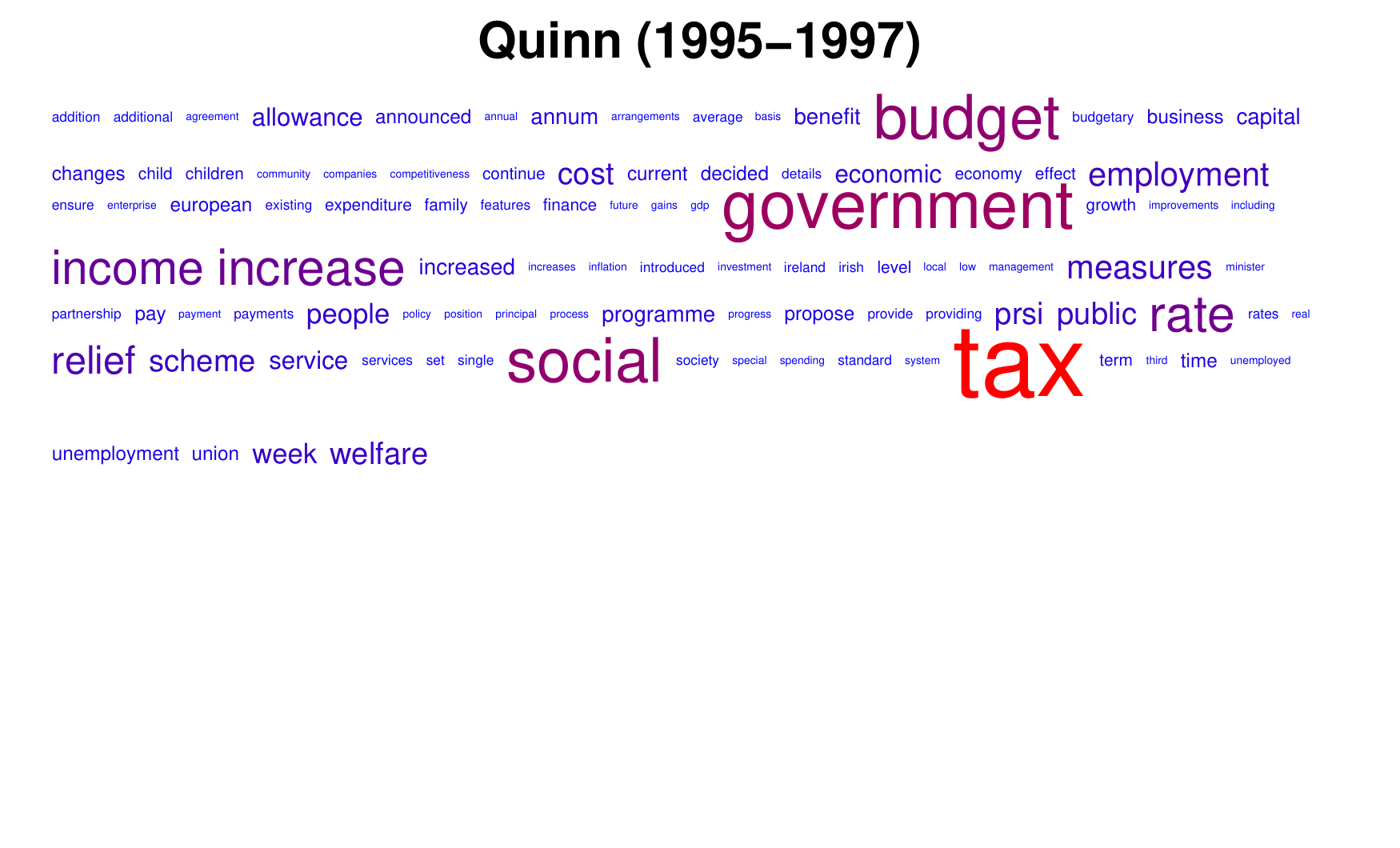}} 
\end{minipage}
\end{center}
\addtocounter{figure}{-1}  
\caption{\emph{Word clouds of all budget speeches made by Ministers for Finance, 1922--2008 (cont'd)}.}
\end{figure}

\begin{figure}
\begin{center}
\begin{minipage}{\textwidth}
\fbox{\includegraphics[width=.49\textwidth]{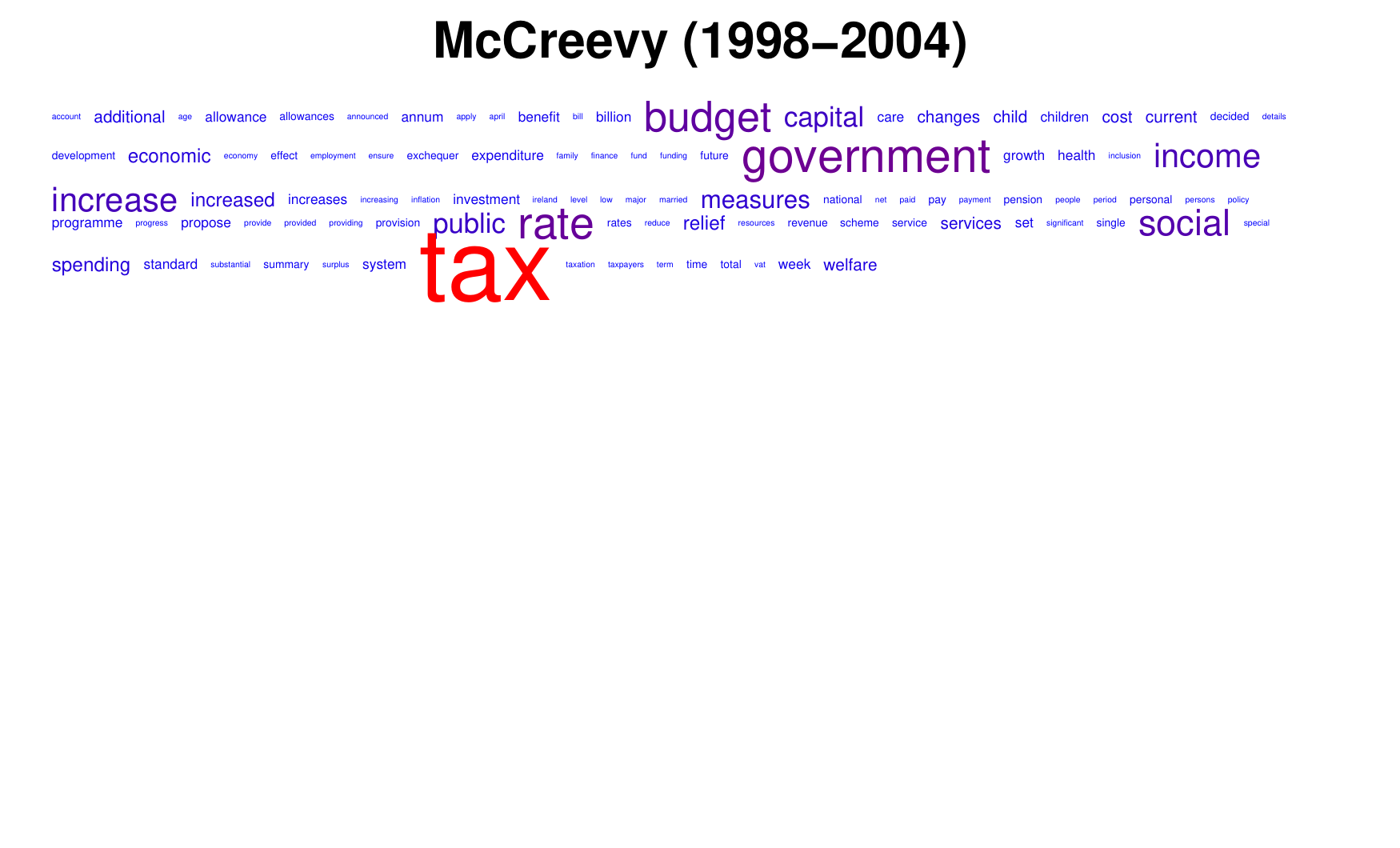}} 
\fbox{\includegraphics[width=.49\textwidth]{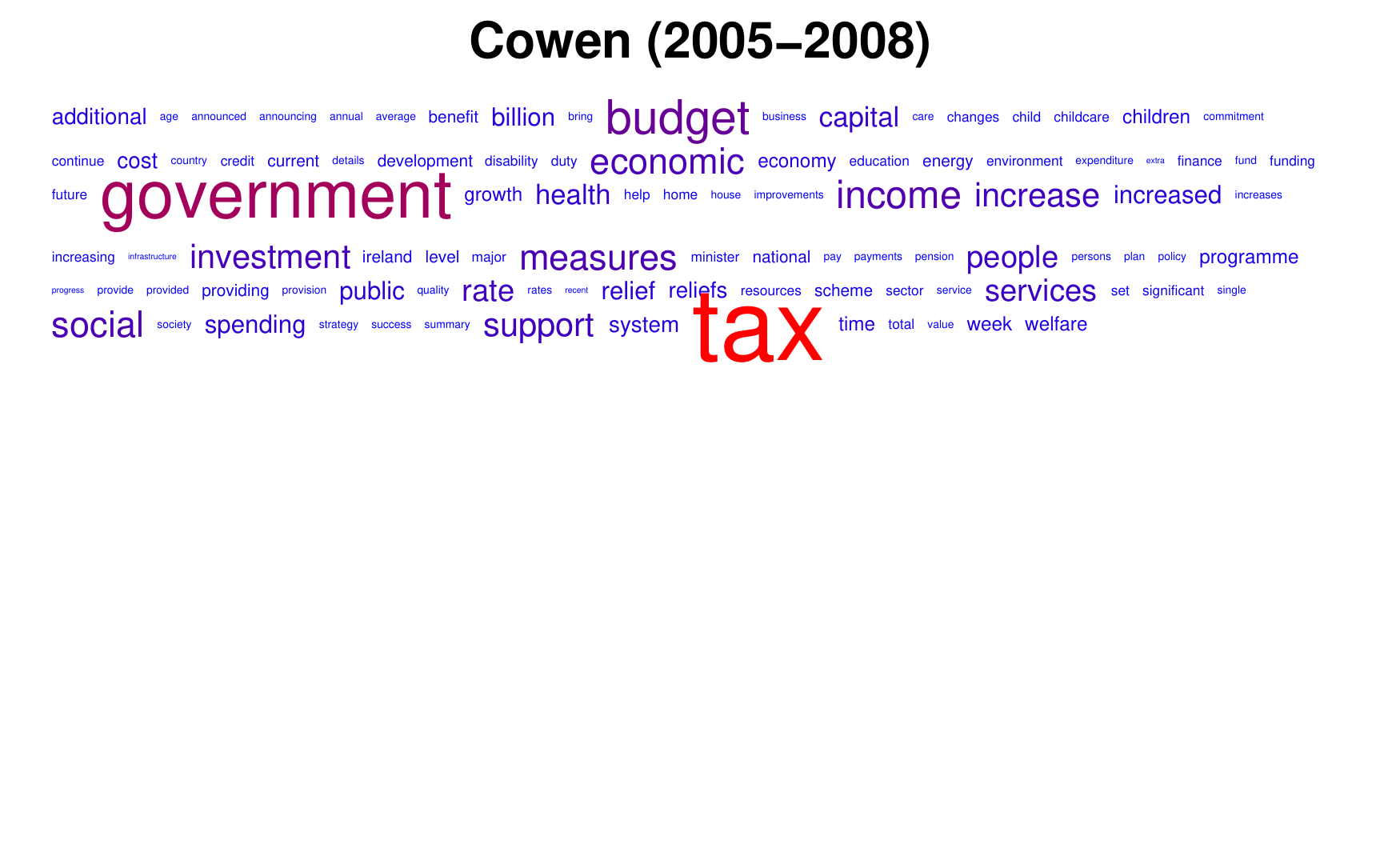}} 
\fbox{\includegraphics[width=.49\textwidth]{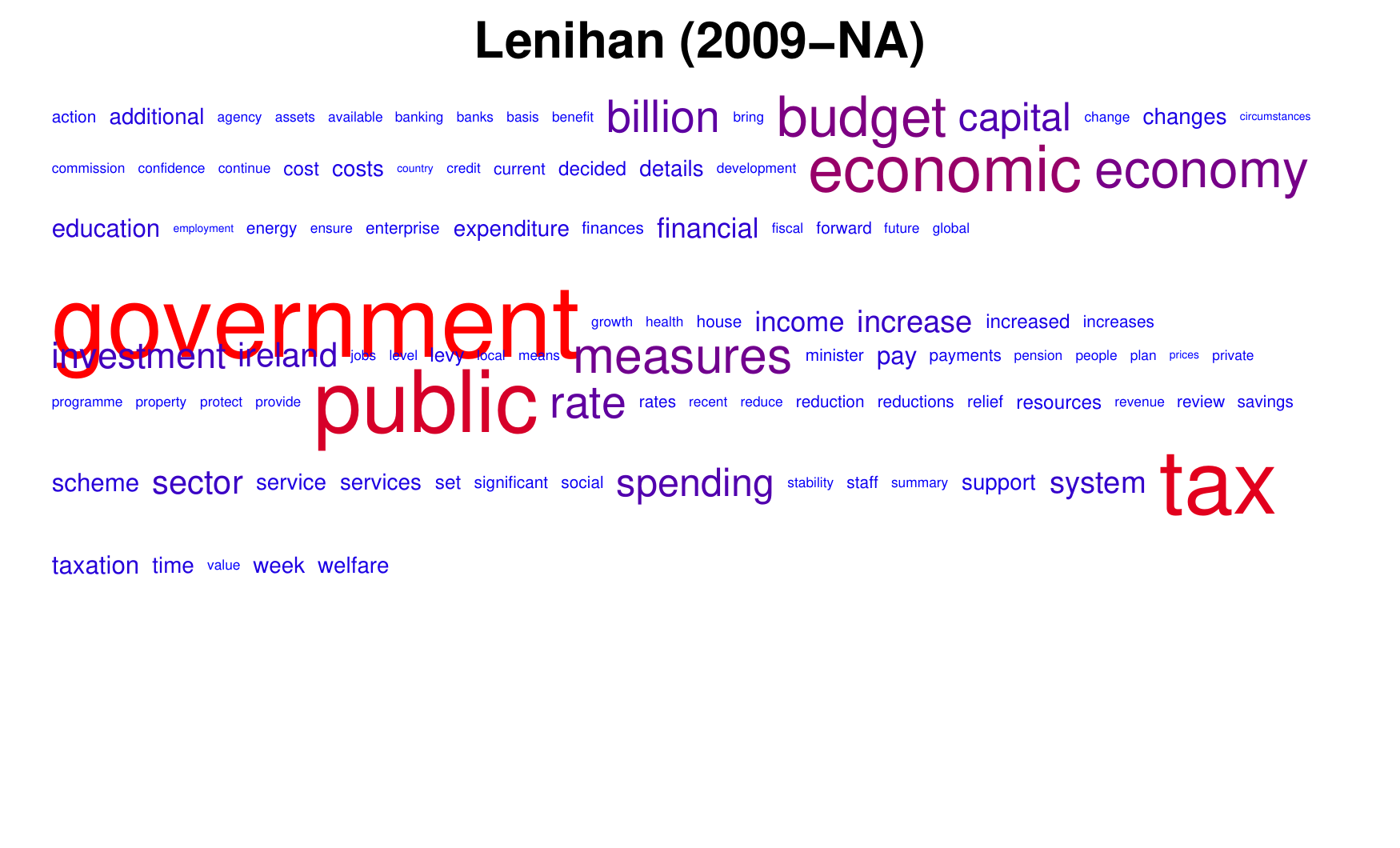}} 
\end{minipage}
\end{center}
\addtocounter{figure}{-1}  
\caption{\emph{Word clouds of all budget speeches made by Ministers for Finance, 1922--2008 (cont'd)}.}
\end{figure}

Overall, while catchy word clouds can only be used as easy first-cut visualizations of the data, rather than methods for any meaningful analysis. One thing that becomes readily apparent from Figure \ref{fig:word_clouds} is that word clouds do not facilitate systematic comparison of documents and their content with one another. Next, we show how our data facilitates the application of relatively simple text analysis techniques to answer more complex empirical questions without the ambiguity in interpretation that is inherent in word clouds.

\subsection{Estimation of Finance Ministers' Policy Positions}
\label{sec:finance_ministers_policy_positions}

Wordfish \citep{Slapin2008} is a method that combines Item Response Theory \citep[e.g.][]{cjr2004} with text classification. Wordfish assumes that there is a latent policy dimension and that each author has a position on this dimension. Words are assumed to be distributed over this dimension such that $y_{ijt} \sim Poisson(\lambda_{ijt})$, where $y_{ijt}$ is the count of word $j$ in document $i$ at time $t$. The functional form of the model is assumed to be

$$
\lambda_{ijt} = \exp(\alpha_{it} + \psi_j + \beta_j \omega_{it})
$$

where $\alpha_{it}$ are fixed effects to control for differences in the length of speeches and $\psi_j$ are fixed effects to control for the fact that some words are used more often than others in all documents. $\omega_{it}$ are the estimates of authors' position on the latent dimension and $\beta_j$ are estimates of word-weights that are determined by how important specific words are in discriminating documents from each other. In this model each document is treated as a separate actor's position and all positions are estimated simultaneously. If a minister maintains a similar position from one budget speech to the next, this means that words with similar frequencies were used over time. At the same time any movement detected by the model towards a position held by, for example, his predecessor, means that the minister's word choice is now much closer to his predecessor than to his own word usage in the previous budget speech. The identification strategy for the model also sets the mean of all positions to 0 and the standard deviation to 1, thus allowing over time a change in positions relative to the mean with the total variance of all positions over time fixed \citep{Slapin2008}. Effectively this standardizes the results and allows for the comparison of positions over time on a comparable scale.

\begin{figure}[htb]
\begin{center}
\fbox{
\includegraphics[width=\textwidth]{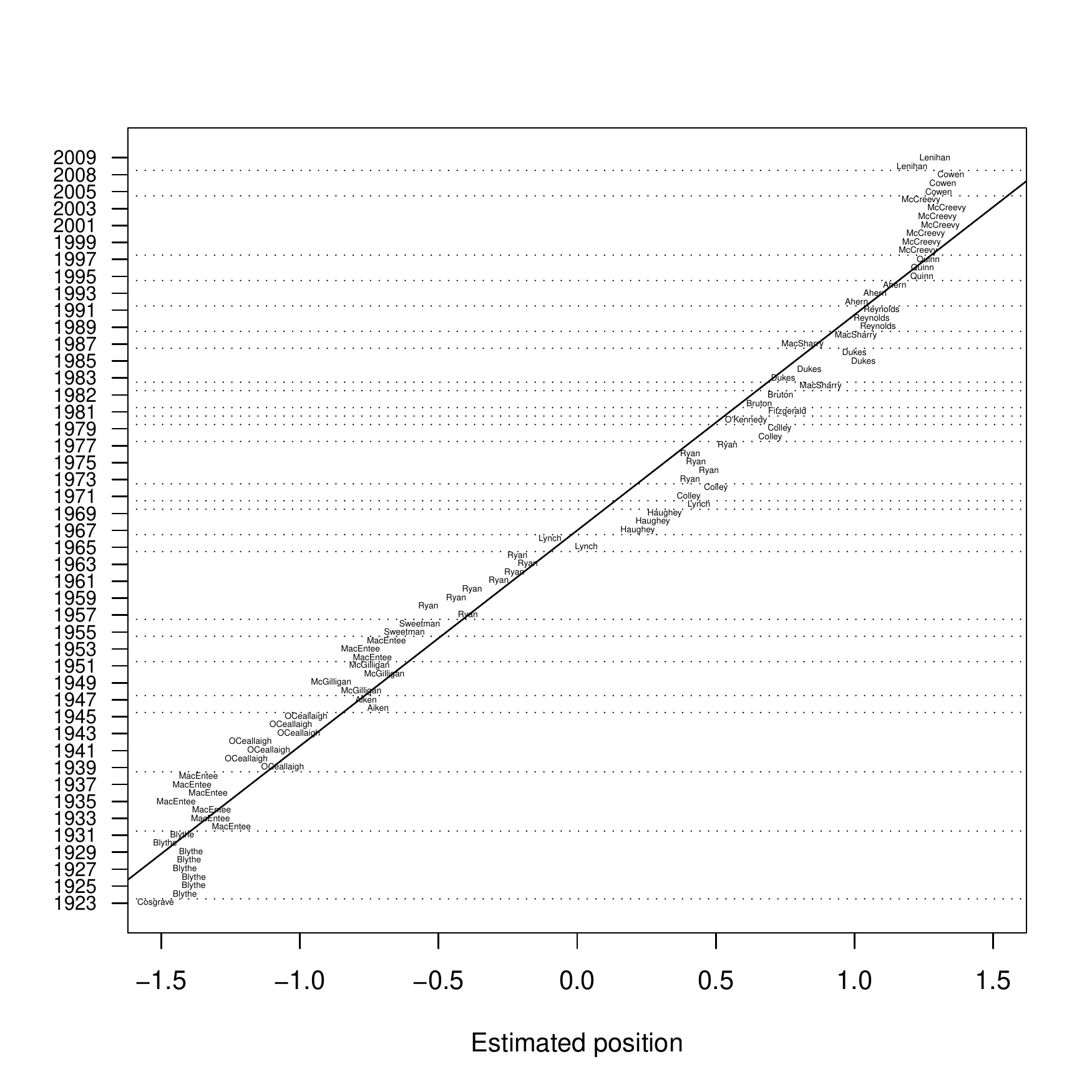}
}
\end{center}
\caption{\emph{Finance ministers' policy positions as estimated from all budget speeches (1922--2009) with an overlaid linear regression line.} 
\label{fig:finance_minister_policy_positions}}
\end{figure}

Before including documents in the analysis, we have removed all numbers, punctuation marks, and stop words. In addition, we follow the advice in \cite{Proksch2009a} and delete words that appear in less than 20\% of all speeches. We do this in order to prevent words that are specific to a small time period (and hence only appear in a few speeches) from having a large impact on discriminating speeches from each other. Figure \ref{fig:finance_minister_policy_positions} shows the results of estimation, with an overlaid regression line. 

The results in Figure \ref{fig:finance_minister_policy_positions} indicate a concept drift -- the gradual change over time of the underlying concept behind the text categorization class \citep[269]{manning2008introduction}. In the political science text scaling literature, this issue is known as agenda shift \citep{Proksch2009a}. In supervised learning models like Wordscore, this problem has typically been dealt with by estimating text models separately for each time period \citep[e.g.][]{baturo2013life,herzog2015}, where the definition of the dimensions remains stable through the choice of training documents. However, this approach is not easily transferrable to inductive techniques like Wordfish, where there may be substantively different policy dimensions at different time periods, rendering comparison of positions over time challenging, if not impossible. A clear presence of the concept drift issue in Wordfish estimation should be a cautionary note for using the approach with time series data, even though the original method was specifically designed to deal with time-series data as indicated in the title of the paper \citep{Slapin2008}. 

Looking at Figure \ref{fig:finance_minister_policy_positions} we can also observe that some ministers have similar preference profiles while others differ significantly. For example, Ahern and Reynolds are very similar in their profile but differ from a group consisting of Quinn, McCreevy, Cowen, and Lenihan who are very close to each other. There also appears to be a dramatic shift in agenda between the tenures of Lynch and Haughey (and also during Taoiseach Lynch's delivery of the budget speech for the Minister for Finance Charles Haughey in 1970). Overall, it appears that topics covered in budget speeches develop in waves, with clear bands formed by, for example, Lenihan, Cowen, McCreevy and Quin; Ahern and Reynolds; MacSharry, Dukes, Bruton, Fitzgerald, O'Kennedy and Colley; R. Ryan, Colley, Lynch (for Haughey); MacEntee, McGilligan and Aiken; Blythe and MacEntee. 

\begin{figure}
\begin{center}
\includegraphics[width=0.49\textwidth]{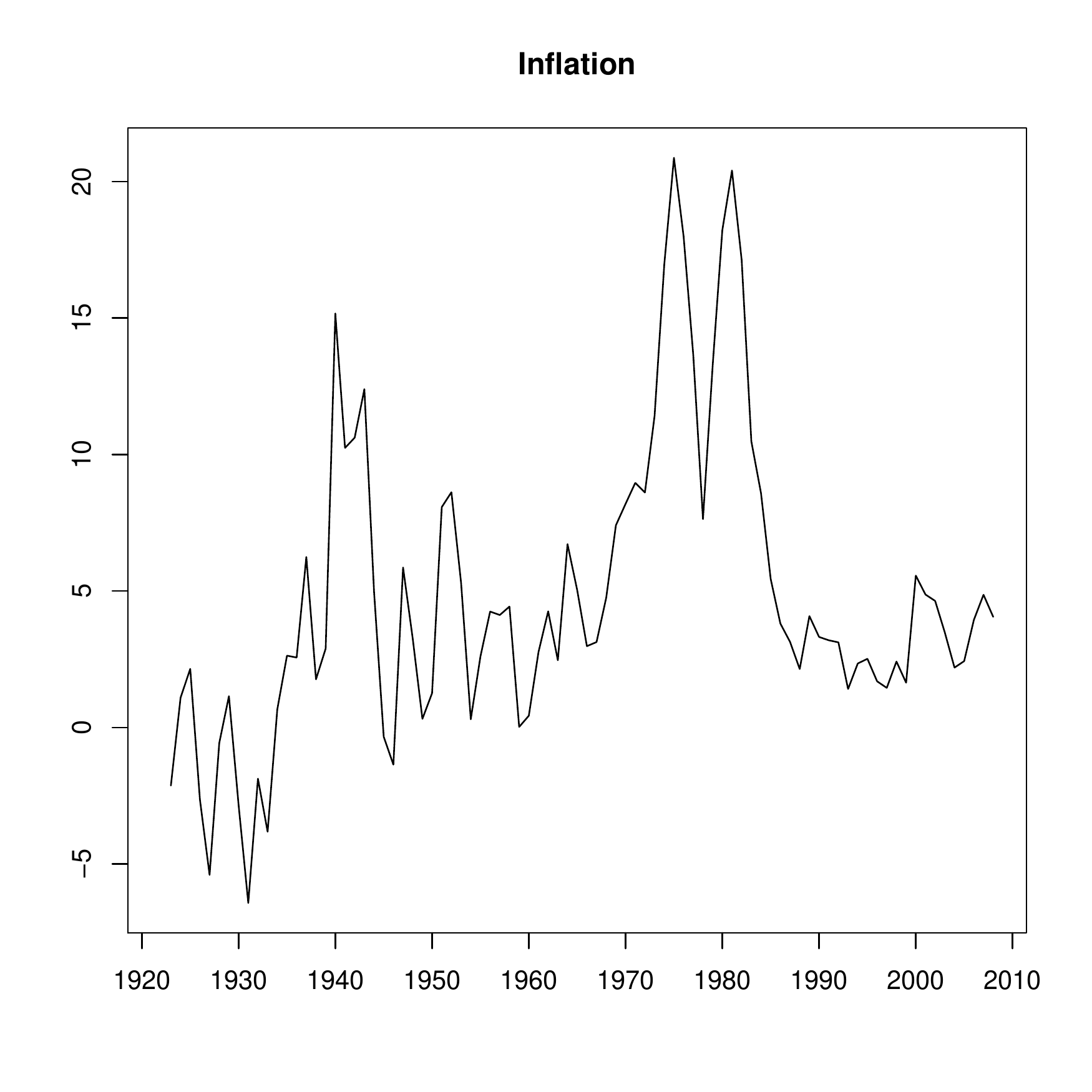}
\includegraphics[width=0.49\textwidth]{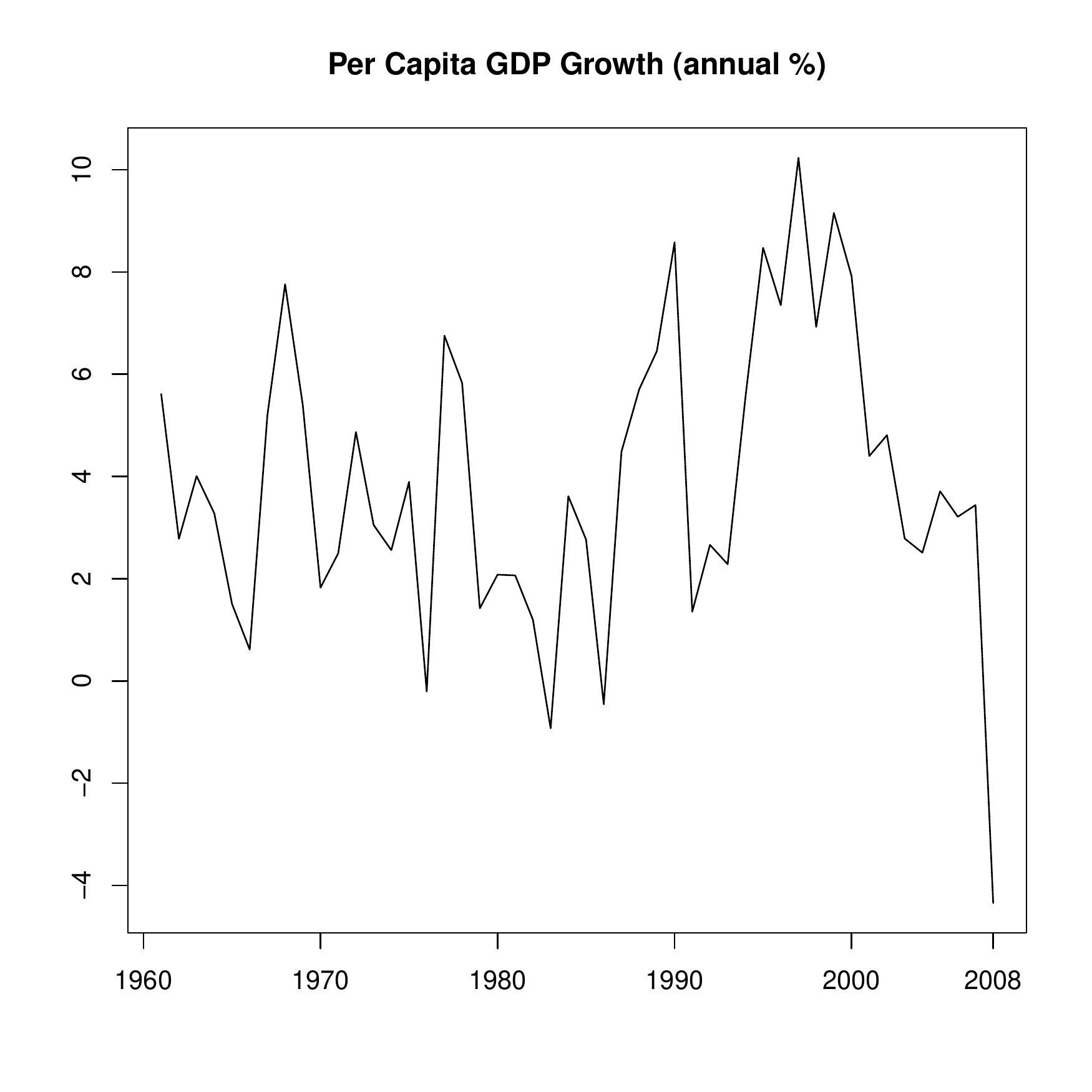}
\includegraphics[width=0.49\textwidth]{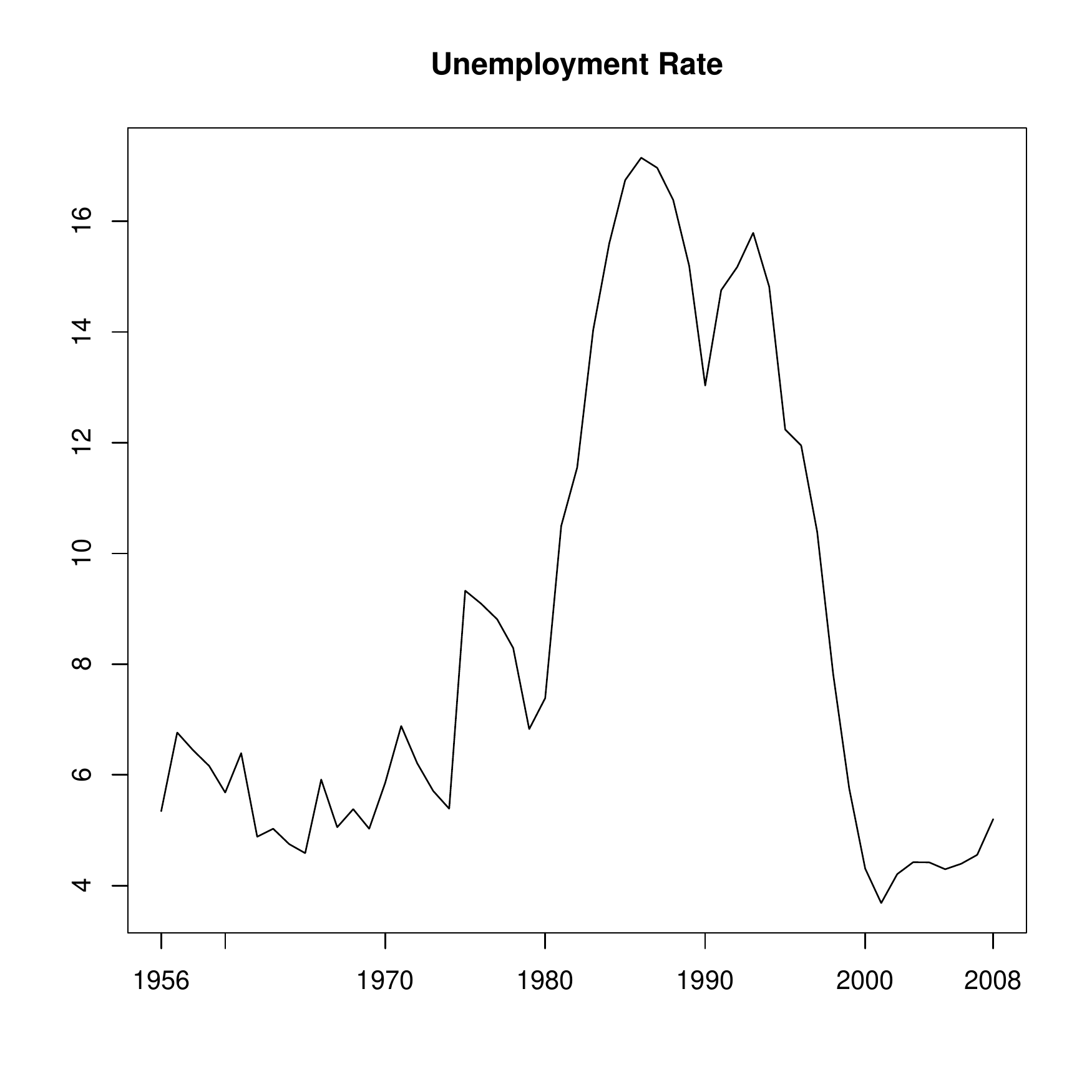}
\end{center}
\caption{\emph{The Irish economy over time: Inflation (1923--2008), Per Capita GDP growth (annual \%; 1961--2008) and unemployment rate (1956--2008).} 
\label{fig:economic_indicators}}
\end{figure}

One intuitive interpretation of our Wordfish results is that budget speeches by finance ministers are related to underlying macroeconomic dynamics in the country. We consider the relationship between estimated policy positions of Minsters and three core economic indicators: unemployment, inflation, and per capita GDP growth rates. Figure \ref{fig:economic_indicators} shows the three economic indicators, inflation (1923--2008), GDP growth (annual \%; 1961--2008) and unemployment rate (1956--2008), over time.

\begin{figure}
\begin{center}
\includegraphics[width=0.49\textwidth]{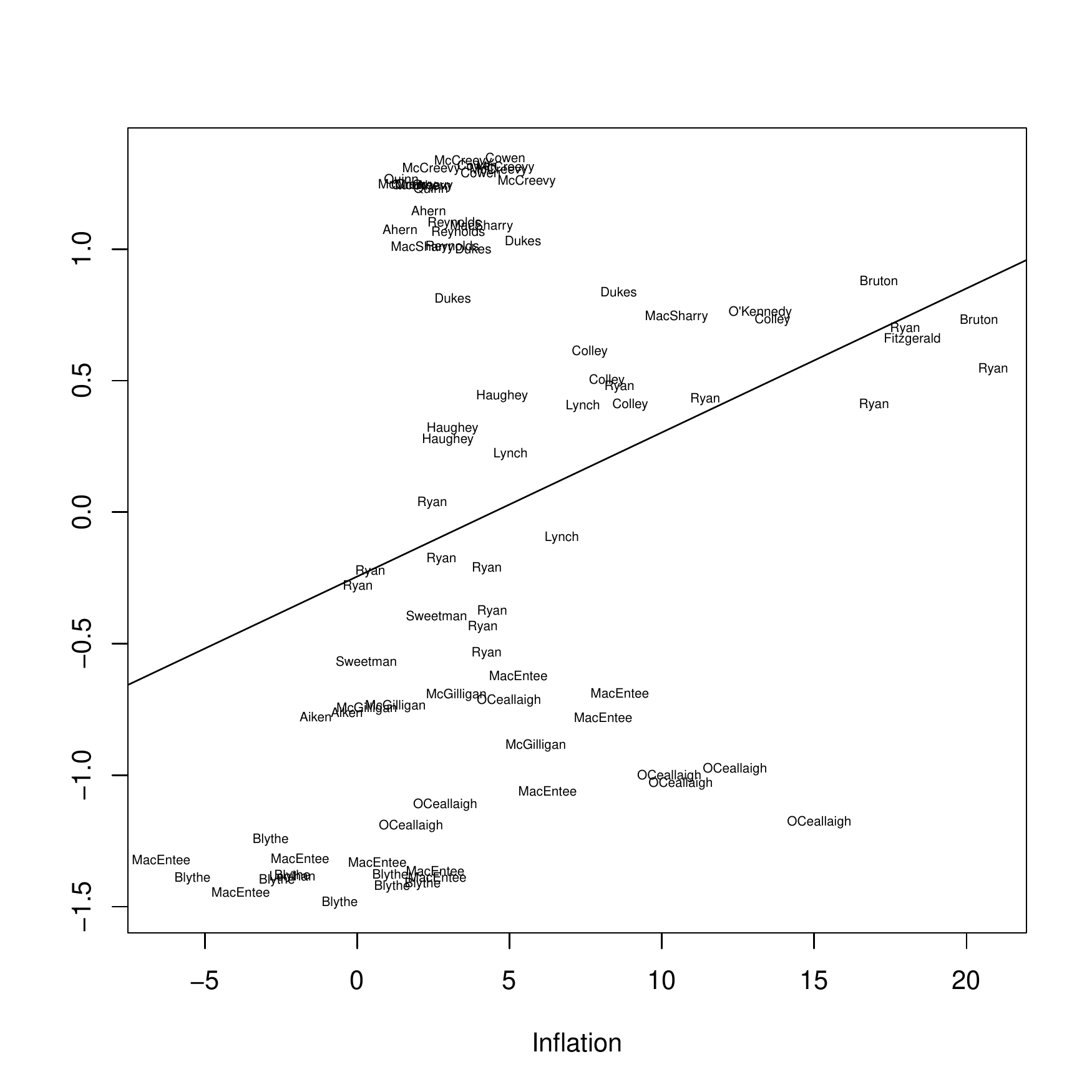}
\includegraphics[width=0.49\textwidth]{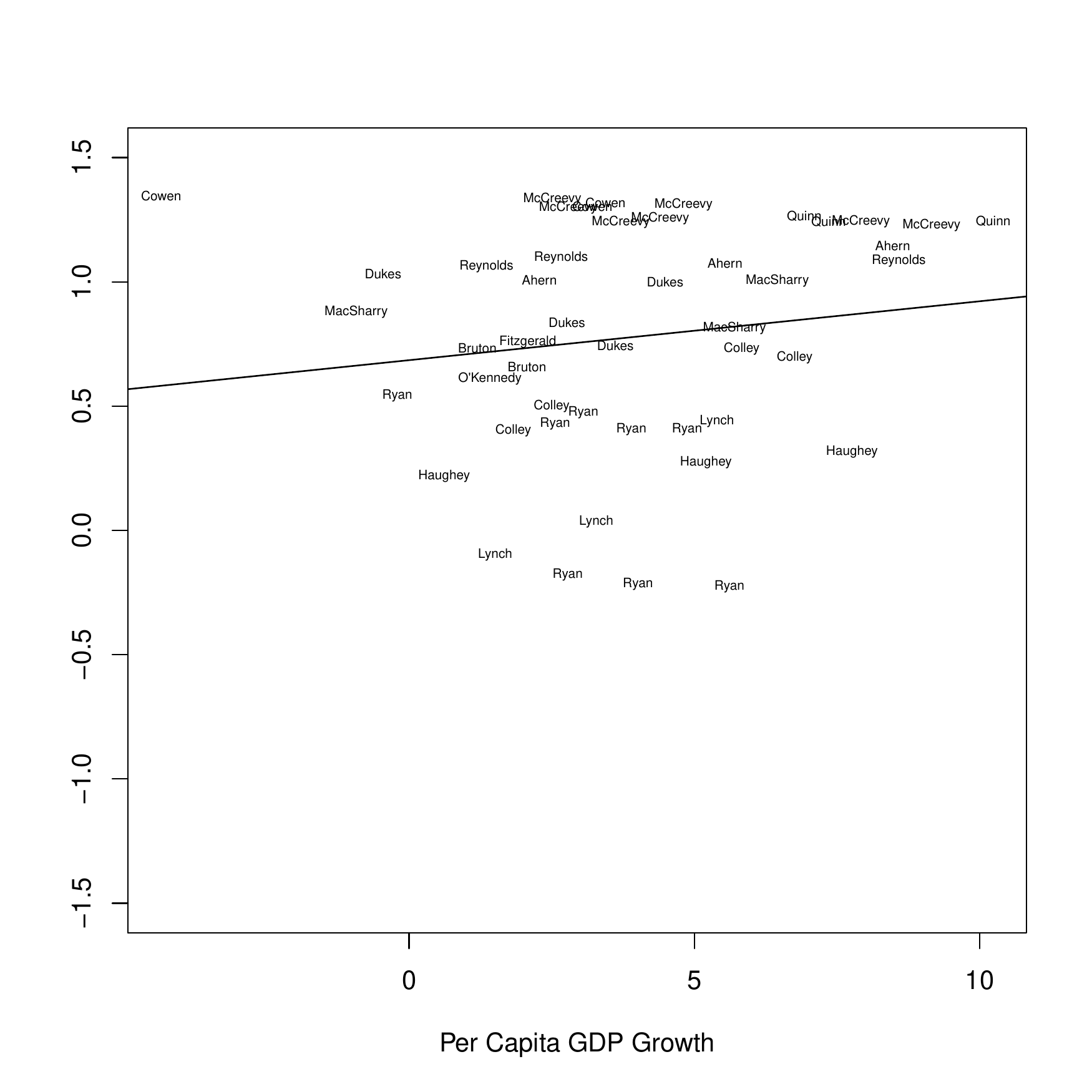}\\
\includegraphics[width=0.49\textwidth]{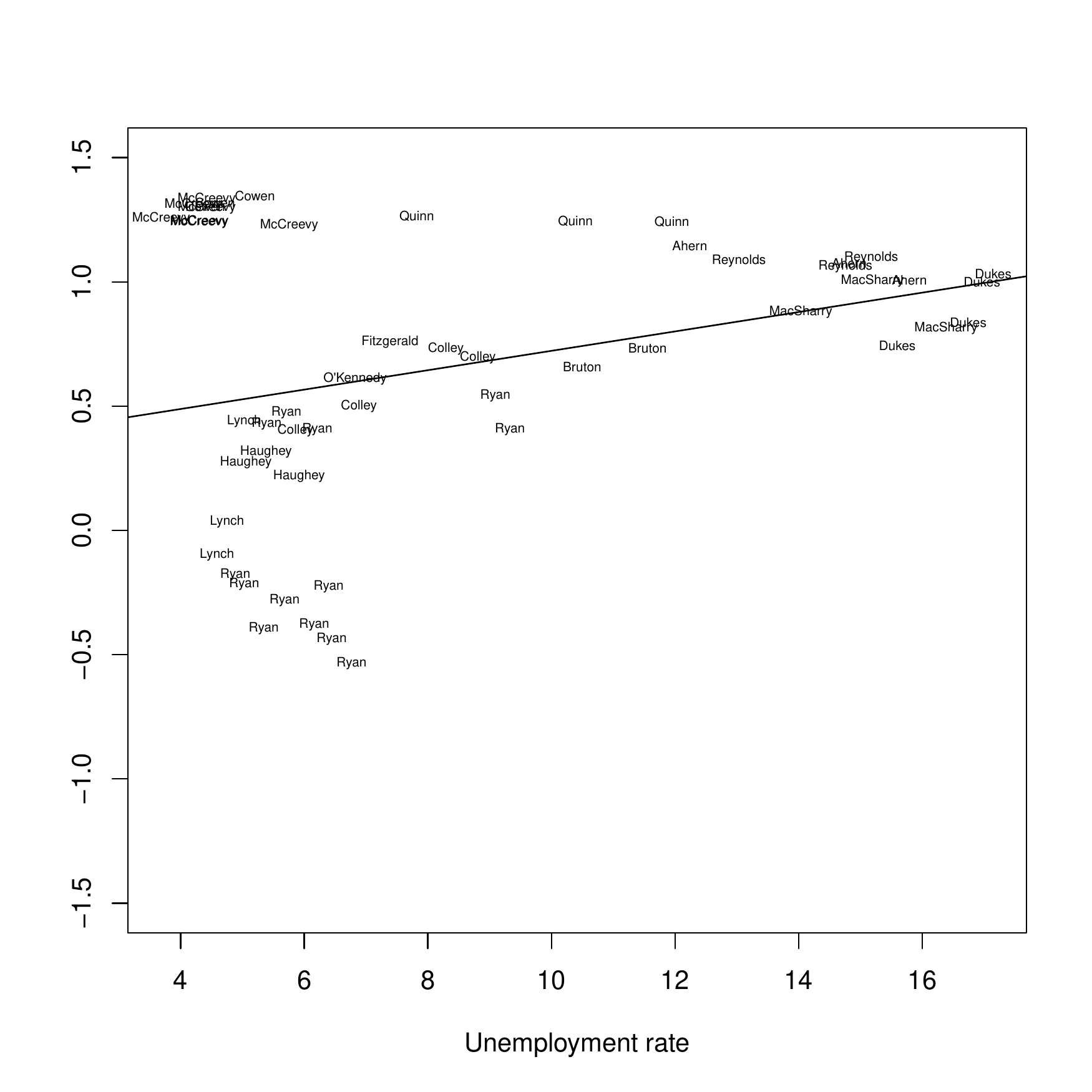}
\end{center}
\caption{\emph{Estimated finance ministers' positions against inflation (1923--2008), GDP growth (annual \%; 1961--2008), and unemployment rate (1956--2008), with an overlaid linear regression line.} 
\label{fig:positions_vs_economy}}
\end{figure}

Figure \ref{fig:positions_vs_economy} show Ministers' estimated positions plotted against the three indicators.        
 
As expected, the results presented here show that the policy positions of some Ministers can be partly explained by the contemporaneous economic situation in the country. However, the fact that some of the Ministers are clear outliers highlights the effect of individual characteristics on policy-making. One of the avenues for research that arises from this exercise is to analyze the determinants of these individual idiosyncrasies, possibly looking at education, class, and previous ministerial career. Such questions can now be easily investigated by researchers using our database.

\subsection{Speakers' Policy Position in the 2008 Budget Debate}
\label{sec:2008_budget_debate}

In the previous section, we used budget speeches from each year and compared them over time. In this section, we restrict the analysis to a single year but take multiple speeches made on the same topic. More specifically, we estimate the preferences of all speakers who participated in the debate over the 2008 budget. We extract these speeches from the database by selecting all contributions to the topic ``Financial Resolution'' in year 2007.\footnote{The debates for the 2008 budget were held in December 2007.} This leaves us with a total of 22 speakers from all five parties. Table \ref{tab:speakers_2008_budget_debate} shows the speeches included in the analysis.

\begin{table}
\caption{Speakers in the 2008 budget debate.\label{tab:speakers_2008_budget_debate}}
\begin{center}
\begin{tabular}{lcC{2cm}C{4cm}}
\toprule
Name & Party & Government party & Length of speech in number of words \\
\midrule
Ahern, Bertie$^1$ & FF & Yes & 3,959 \\
Ahern, Dermot & FF & Yes & 2,700 \\
Ahern, Michael & FF & Yes & 1,190 \\
Ardagh, Sean & FF & Yes & 1,015 \\
Carey, Pat & FF & Yes & 942 \\
Cowen, Brian$^2$& FF & Yes & 8,733 \\
Dempsey, Noel & FF & Yes & 1,438\\
Devins, Jimmy & FF & Yes & 1,090 \\
O'Keeffe, Batt & FF & Yes & 715 \\[.2cm]
Gormley, John & Green & Yes & 4,306 \\[.2cm]
Bruton, Richard & FG & No & 10,817\\
Burke, Ulick & FG & No & 714 \\
Hogan, Phil & FG & No & 1,438 \\
Kenny, Enda$^3$ & FG & No & 3,924 \\
Neville, Dan & FG & No & 1,210 \\
O'Donnell, Kieran & FG & No & 1,182 \\
Reilly, James & FG & No & 1,683 \\
Varadkar Leo & FG & No & 1,876 \\[.2cm]
Gilmore, Eamon & Labour & No & 5,141 \\
Shortall, Roisin & Labour & No & 2,662 \\[.2cm]
Morgan, Arthur & SF & No & 6,158 \\
O'Caolain, Caoimhghin & SF & No & 1,438 \\
\bottomrule
\end{tabular}
\vspace{0.2cm}
\begin{small}
\begin{minipage}{13cm}
\emph{Notes}: 1--Taoiseach, 2--Minister for Finance, 3--FG Party Leader. The budget debate for the 2008 budget was held in December 2007.
\end{minipage}
\end{small}
\end{center}
\end{table}

To estimate speakers' position we use Wordscore \citep{Laver2003} -- a version of the Naive Bayes classifier that is deployed for text categorization problems \citep{benoitnulty:2013}. In a similar application, \cite{Laver2003} have already demonstrated that Wordscore can be effectively used to derive estimates of TDs policy positions.\footnote{\cite{Laver2002} used Wordscore to estimate TDs position in the 1991 confidence debate on the future of the Fianna F{\'a}il--PD coalition government.} As in the example above, we pre-process documents by removing all numbers and interjections. 

Wordscore uses two documents with well-known positions as reference texts (training set). The positions of all other documents are then estimated by comparing them to these reference documents. The underlying idea is that a document that, in terms of word frequencies, is similar to a reference document was produced by an author with similar preferences. The selection of reference documents furthermore determines the (assumed) underlying dimension for which documents' positions are estimated. For example, using two opposing documents on climate change would scale documents on the underlying dimension ``climate politics''. It has also been shown that under certain assumptions the Wordscore algorithm is related to the Wordfish algorithm used in the previous section \citep{Lowe2008}. 

\begin{figure}
\begin{center}
  \includegraphics[width=\textwidth]{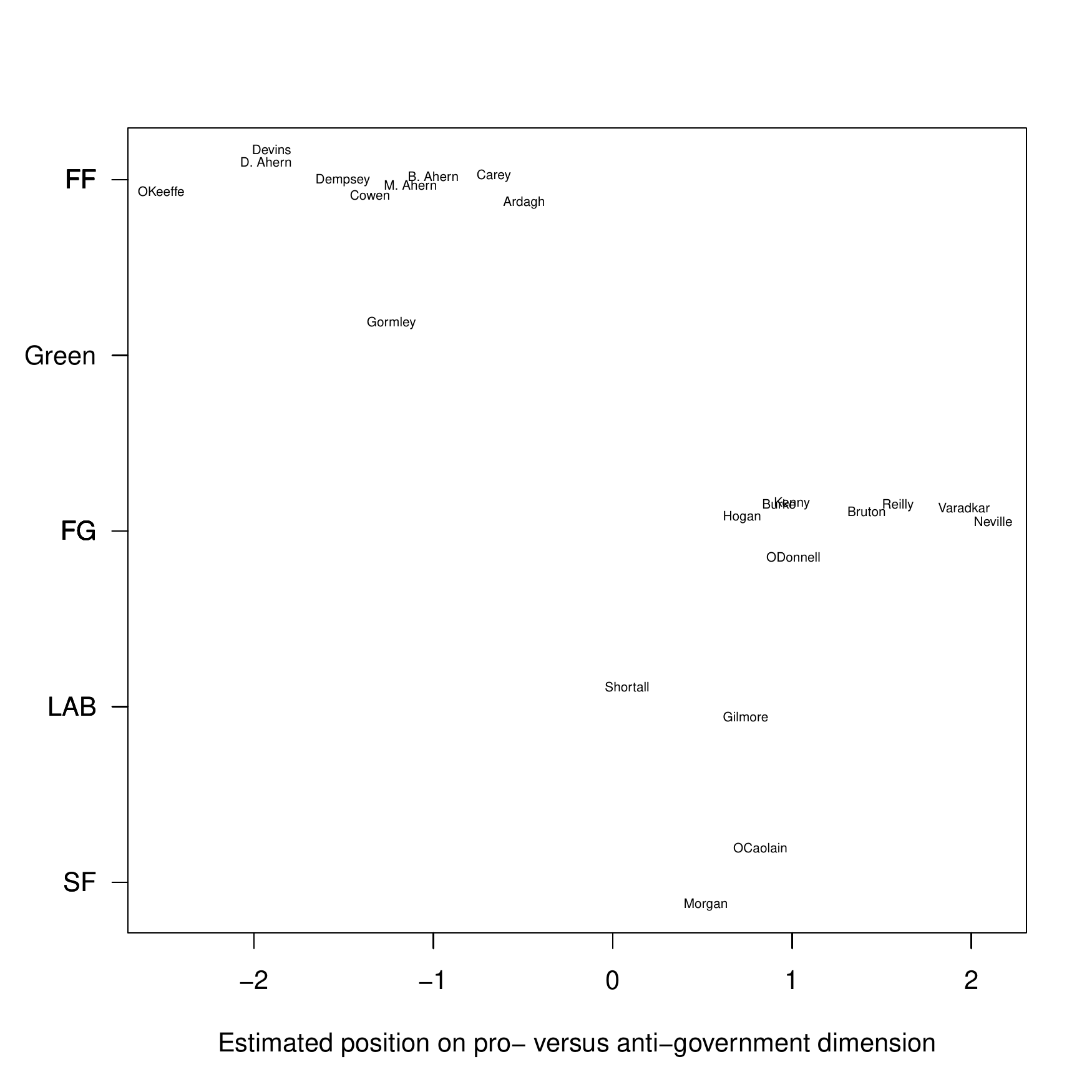}
\end{center}
\caption{\emph{Estimated positions of all speakers in the 2008 budget debate. Estimated dimension represents pro- versus anti-government positions. Scaling of x-axis is arbitrary. Speeches of Bertie Ahern (FF, Taoiseach) and Enda Kenny (FG party leader) were used as reference texts for being respectively pro- or anti-government. Observations are jittered along the y-axis to prevent names from overlapping.}}
\label{fig:budget_debate_2008}
\end{figure}

We assume that contributions in budget debates have the underlying dimension of being either \emph{pro} or \emph{contra} the current government. Our interpretation from reading the speeches is that, apart from the budget speech itself, all other speeches largely either attack or defend the incumbent government and to a lesser extent debate the issues of the next budget. We can therefore use contributions during the budget debate as an indicator for how much a speaker is supporting or opposing the current government, here consisting of Fianna F{\'a}il and the Green Party. As our reference texts we therefore chose the speeches of Bertie Ahern (Taoiseach) and Enda Kenny (FG party leader). The former should obviously be strongly supportive of the government while the latter, as party leader of the largest opposition party, should strongly oppose it. Figure \ref{fig:budget_debate_2008} shows estimated positions for all speakers grouped by party affiliation.

The estimated positions are clustered into two groups, one representing the government and one the opposition. Within the government cluster, Deputy Batt O'Keeffe (Minister of State at the Department of Environment, Heritage and Local Government) is estimated to be the most supportive speaker for the government, while Deputy Pat Carey (Minister of State at the Department of Community, Rural and Gaeltacht Affairs) and Deputy Sean Ardagh are estimated to be relatively closer to the opposition. Deputy John Gormley, leader of the Green party and Minister for the Environment, Heritage and Local Government in the FF-Green coalition, is estimated to be in the centre of the government cluster. Among all positions in the opposition cluster, the speech of R{\'o}is{\'i}n Shortall is the closest to the government side, with Neville being the farthest out. 

\subsection{Ministers' policy position in the 26th government}
\label{sec:ministers_policy_position_in_the_26th_government}

The government cabinet in parliamentary democracies is at the core of political decision making, yet it is difficult to model intra-cabinet bargaining as the preferences of most cabinet members are unknown.  Cabinet decisions are usually made behind closed doors and the doctrine of joint cabinet responsibility prevents ministers from publicly opposing decisions, even if they disagree with them. Using ministers' speeches and their responses during question times offer a unique opportunity to infer their preferences on policy dimensions of interest. In our final application we estimate policy positions for all cabinet members in the 26th government. The dimension on which positions are estimated represents pro- versus contra-government spending (or spending left-right). We show that estimated positions are highly correlated with departments' actual spending, which means that estimated positions are not only meaningful but can also be used to predict actual policy-making. 

The 26th government was formed as a coalition between Fianna F{\'a}il and the Progressive Democrats after the election for the 29th D{\'a}il in 2002. The cabinet was reshuffled on 29 September 2004 and we only include ministers' speeches until that date. Table \ref{tab:ministers_26th_government} lists all cabinet members (and their portfolios) included in our analysis.

\begin{table}
\begin{center}
\caption{Members of the 26th government (29th D{\'a}il), 6 June 2002--29 September 2004 \label{tab:ministers_26th_government}}
\begin{tabular}{lC{0.1\textwidth}L{0.6\textwidth}}
\toprule
\textbf{Name} & \textbf{Party} & \textbf{Office} \\
\midrule
Bertie Ahern             & FF  & Taoiseach \\
Mary Harney              & PD  & T{\'a}naiste and Minister for Enterprise, Trade and Employment \\
Michael Smith            & FF  & Minister for Defence \\
Joe Walsh                & FF  & Minister for Agriculture and Food \\
Charlie McCreevy         & FF  & Minister for Finance \\
Brian Cowen              & FF  & Minister for Foreign Affairs \\
Noel Dempsey             & FF  & Minister for Education and Science \\
Dermot Ahern             & FF  & Minister for Communications, Marine and Natural Resources \\
John O'Donoghue          & FF  & Minister for Arts, Sport and Tourism \\
Miche{\'a}l Martin       & FF  & Minister for Health and Children \\
S{\'e}amus Brennan       & FF  & Minister for Transport \\
Michael McDowell         & PD  & Minister for Justice, Equality and Law Reform \\
Martin Cullen            & FF  & Minster for the Environment and Local Government \\
{\'E}amon {\'O} Cu{\'i}v & FF  & Minister for Community, Rural and Gaeltacht Affairs \\
Mary Coughlan            & FF  & Minister for Social and Family Affairs \\
\bottomrule
\multicolumn{3}{l}{\small{\emph{Source:} Houses of the Oireachtas (http://www.oireachtas.ie/viewdoc.asp?DocID=2935).}}
\end{tabular}
\end{center}
\end{table}

To estimate ministers' policy positions, we retrieve the complete record of each minister's contribution in parliament from the first meeting on 6 June 2002 until the date of the reshuffle. On average, each minister made 3,643 contributions with an average number of 587,077 words. Table \ref{tab:word_count_minister_speeches} provides summary statistics for all ministers, sorted by total word count.

\begin{table}
\begin{center}
\caption{Summary statistics for ministers' contributions in the 26th government, 6 June 2002 -- 29 September 2004, sorted by total word count \label{tab:word_count_minister_speeches}}
\begin{tabular}{L{0.3\textwidth}C{0.2\textwidth}C{0.2\textwidth}C{0.2\textwidth}}
\toprule
\textbf{Name} & \textbf{Party} & \textbf{Number of contributions} & \textbf{Total word count} \\
\midrule
Noel Dempsey             & FF & 8,066 & 1,273,835 \\ 
Michael McDowell         & PD & 6,290 & 1,038,527 \\ 
Bertie Ahern             & FF & 6,505 &   790,964 \\ 
Dermot Ahern             & FF & 3,047 &   755,471 \\ 
Charlie McCreevy         & FF & 3,249 &   657,010 \\ 
Brian Cowen              & FF & 2,444 &   652,062 \\ 
Martin Cullen            & FF & 5,826 &   574,464 \\ 
S{\'e}amus Brennan       & FF & 3,324 &   513,938 \\ 
Mary Coughlan            & FF & 2,627 &   503,413 \\ 
Mary Harney              & PD & 3,357 &   418,745 \\ 
Michael Smith            & FF & 2,464 &   330,575 \\ 
{\'E}amon {\'O} Cu{\'i}v & FF & 1,459 &   286,194 \\ 
John O'Donoghue          & FF & 1,553 &   282,154 \\ 
Miche{\'a}l Martin       & FF &   789 &   141,721 \\ 
\bottomrule
\multicolumn{4}{l}{\small{\emph{Note:} Only speeches before the cabinet reshuffle on 29 September 2004 are included.}}
\end{tabular}
\end{center}
\end{table}

\begin{figure}
\begin{center}
  \includegraphics[width=\textwidth]{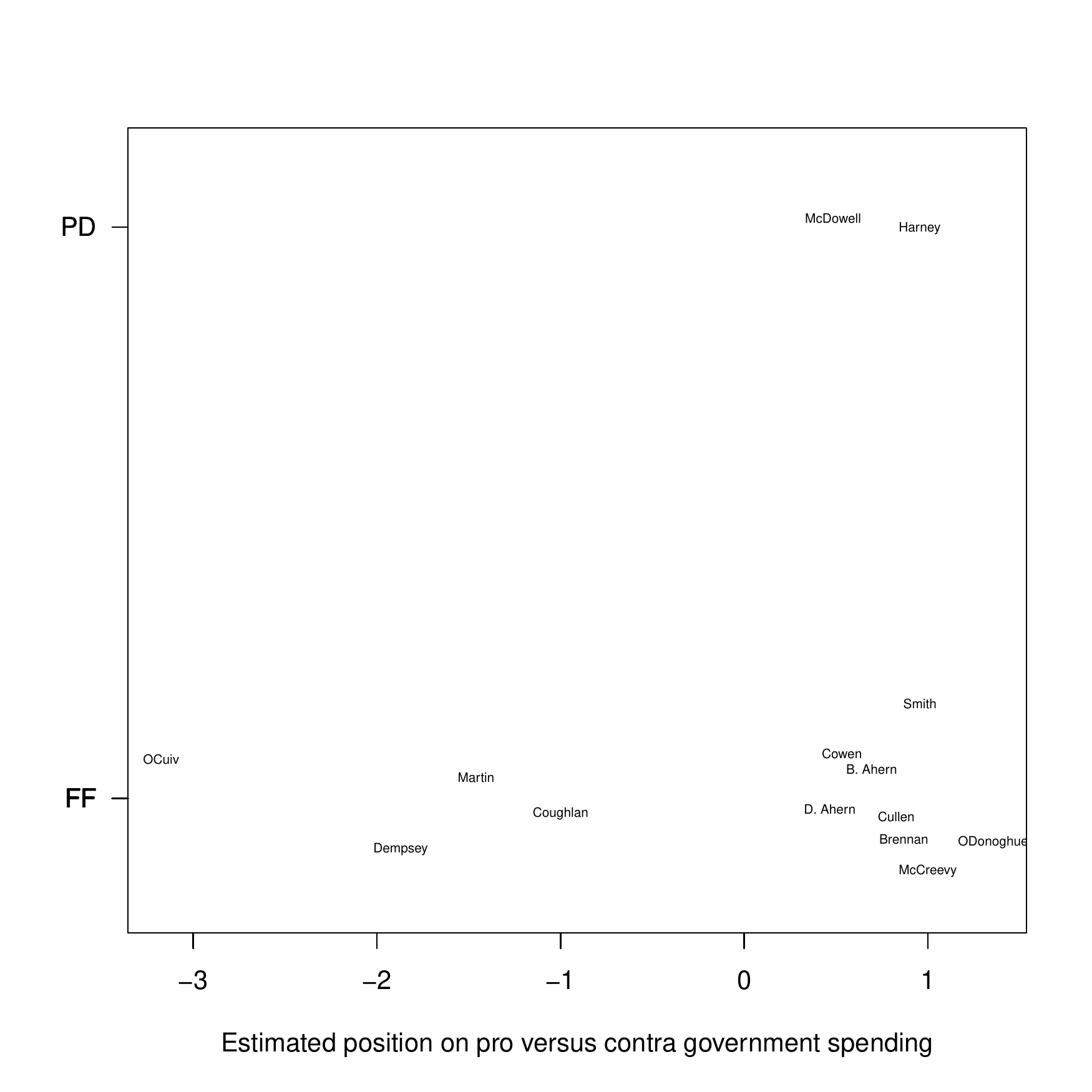}
\end{center}
\caption{\emph{Estimated positions for all cabinet members in the 26th government (29th D{\'a}il) using Wordscore. Positions are jittered along the y-axis. Estimation is based on each minister's contribution in D{\'a}il {\'E}ireann before the cabinet reshuffle on 29 September 2004. Speeches by Mary Coughlan (Minister for Social and Family Affairs) and Charlie McCreevy (Minister for Finance) are used as left and right reference texts, respectively}}
\label{fig:minister_positions_dail29}
\end{figure}

We again use Wordscore \citep{Benoit2003,Laver2003} to estimate positions as it allows us to define the underlying policy dimension by choosing appropriate reference texts. We estimate positions on a social-economic left-right dimension that reflects pro- versus contra-government spending. We therefore use contributions by Mary Coughlan (Minister for Social and Family Affairs) and Charlie McCreevy (Minister for Finance) as reference texts, assuming that the former is more in favor of spending than the latter. Figure \ref{fig:minister_positions_dail29} shows the results of estimation grouped by the two parties. 

As expected, we find that the two PD members, Mary Harney and Michael McDowell, are at the right side of the dimension. We estimate the most left-wing members to be {\'E}amon {\'O} Cu{\'i}v (Minister for Community, Rural and Gaeltacht Affairs), Noel Dempsey (Minister for Education and Science), and Miche{\'a}l Martin (Minister for Health and Children). The most right-wing members are John O'Donoghue (Minister for Arts, Sport and Tourism), Charlie McCreevy (whose contributions we used as right-wing reference text), and Michael Smith (Minister for Defense).

\begin{figure}
\begin{center}
\includegraphics[width=.45\textwidth]{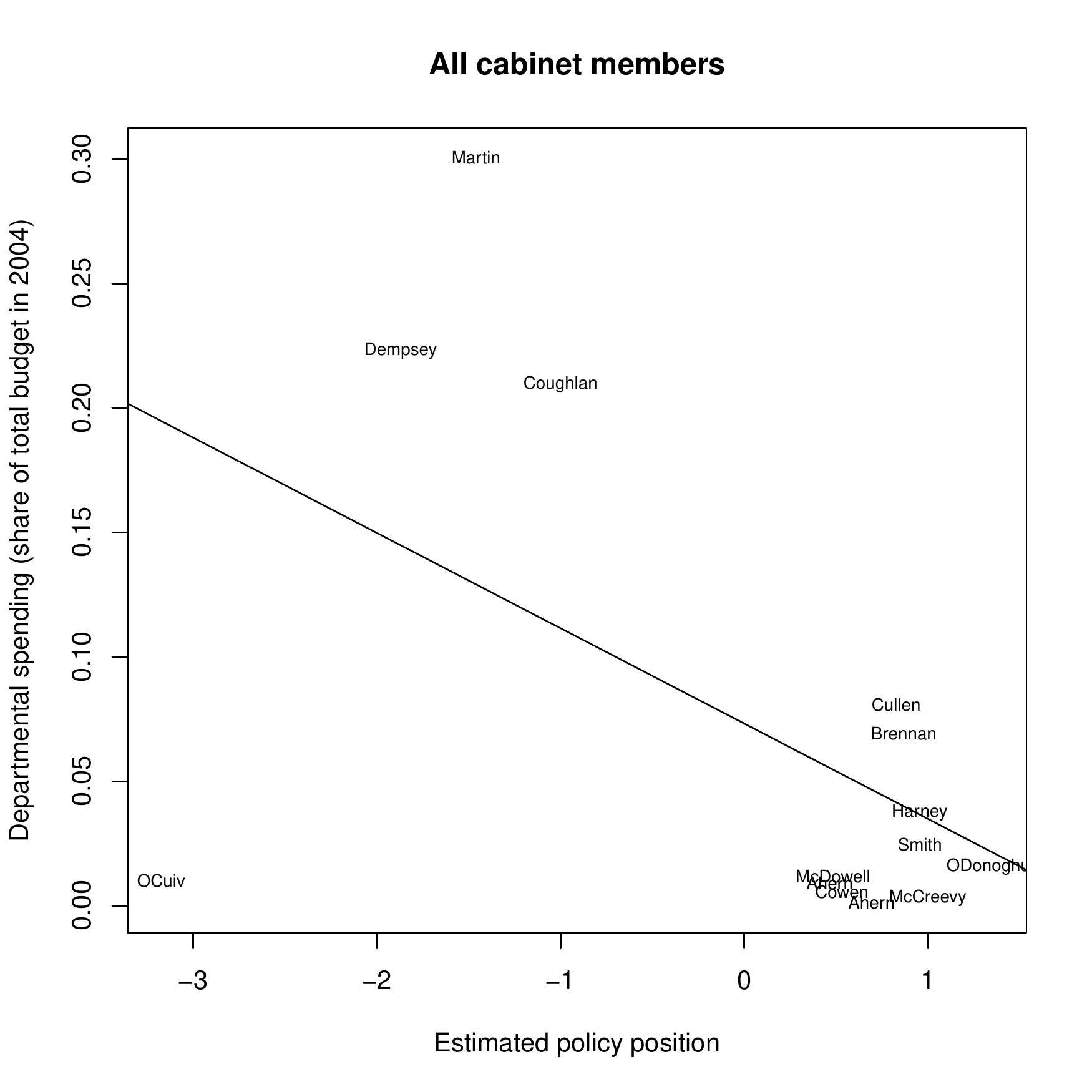}
\includegraphics[width=.45\textwidth]{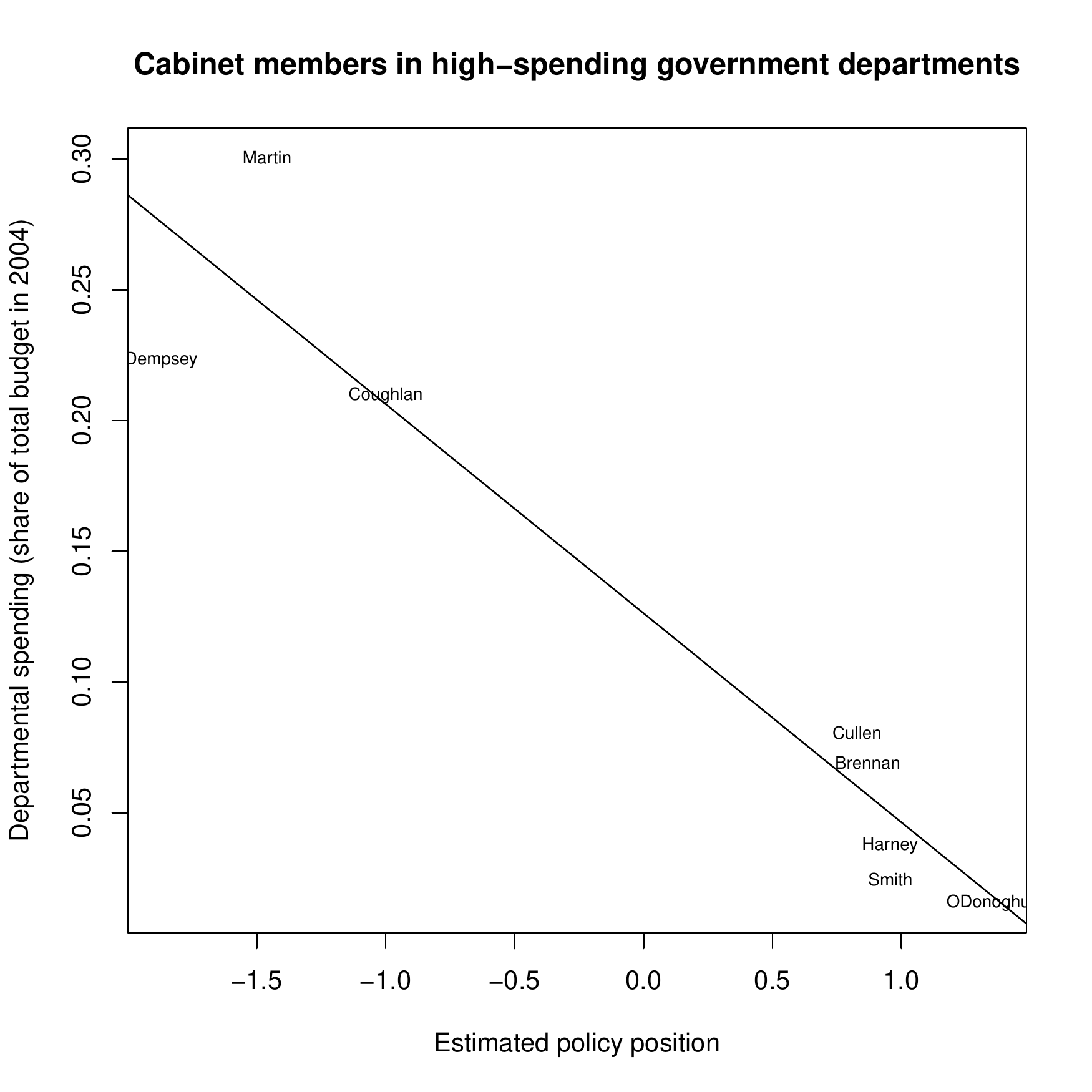}
\end{center}
\caption{\emph{Cabinet ministers' policy position plotted departmental spending as share of total government budget in 2004. For the analysis of high-spending departments we remove the Office of the Taoiseach or the Department of Foreign Affairs, with the remaining eight departments accounting for more than 95 per cent of the total budget in 2004 }}
\label{fig:minister_positions_vs_budget_share_all_ministers}
\end{figure}

How valid are these estimated positions? In order to have substantive meaning, our estimates should be able to predict political decisions on the same policy dimension. We therefore use ministers' estimated positions to predict their departmental spending level \citep[see][for a similar analysis with data from Italy]{Giannetti2005}. Our outcome variable is each department's spending as share of the total budget in 2004 modeled as a function of estimated policy positions. We conjecture that more left-wing ministers should have higher spending levels than right-wing ministers, which we test by estimating

\begin{equation}
\label{eq:spending_regression}
\widehat{\text{spending}} = \hat{\beta}_0 + \hat{\beta}_1 \text{policy position}
\end{equation}

via ordinary least-square regression. Figure \ref{fig:minister_positions_vs_budget_share_all_ministers} shows the two variables plotted against each other together with the estimated regression line from equation \ref{eq:spending_regression}. In one analysis shown we include all cabinet members. In the other, we exclude non-spending departments with small budgets, such as the office of the Taoiseach or the Department of Foreign Affairs.\footnote{The eight high-spending departments we include are, in decreasing order of budget share, the Department of Health and Children, Department of Education and Science, Department of Social and Family Affairs, Department of the Environment and Local Government, Department of Transport, Department of Enterprise, Trade and Employment, Department of Defence, and the Department of Arts, Sport and Tourism. These eight departments together account for more than 95 per cent of the total budget in 2004.}

Figure \ref{fig:minister_positions_vs_budget_share_all_ministers} reveals that there is a negative, albeit weak, relationship between estimated positions and spending, with more left-wing cabinet members having higher spending levels than right-wing members. The correlation between the two variables is -0.53 ($p = 0.0523$) which is not significant at the 0.05 level. However, if we only take members from high-spending departments into account (second pane in Figure \ref{fig:minister_positions_vs_budget_share_all_ministers}) we find a significant linear relationship between the two variables with a correlation coefficient of -0.95 ($p = 0.0002$). This result provides some level of validation for our data and analysis. 

These results also open up an intriguing question about the endogeneity of observable policy preferences of ministers. Do higher spending portfolios receive more pro-spending ministers or do ministers adapt their policy preferences after appointment and literally grow into the job? This and related questions are outside the scope of this paper and can be pursued by researchers with the help of our database of parliamentary speeches.

\section{Conclusion}

Policy preferences of individual politicians (ministers or TDs in general), are inherently unobservable. However, we have abundant data on speeches made by political actors. The latest developments in automated text analysis techniques allow us to estimate the policy positions of individual actors from these speeches. 

In relation to Irish political actors such estimation has been hindered by the structure of the available data. While all speeches made in D{\'a}il {\'E}ireann are dutifully recorded, the architecture of the data set, where digitized versions of speeches are stored, makes it impossible to apply any of the existing text analysis software. Speeches are currently stored by D{\'a}il {\'E}ireann in more than half a million separate HTML files with entries that are not related to one another. 

In this paper we present a new database of speeches that was created with the purpose of allowing the estimation of policy preferences of individual politicians. For that reason we created a relational database where speeches are related to the members database and structured in terms of dates, topics of debates, and names of speakers, their constituency and party affiliation. This gives the necessary flexibility to use available text scaling methods in order to estimate the policy positions of actors. 

We also present several examples for which this data can be used. We show how to estimate the policy positions of all Irish Ministers for Finance, and highlight how this can lead to interesting research questions in estimating the determinants of their positions. We show that for some ministers the position can be explained by the country's economic performance, while the preferences of other ministers seem to be idiosyncratic. In another example we estimate positions of individual TDs in a budget debate, followed by the estimation of policy positions of cabinet members of the 26th Government. 

With the introduction of our database, we aim to make text analysis an easy and accessible tool for social scientists engaged in empirical research on policy-making that requires estimation of policy preferences of political actors.

\newpage \singlespacing
 \bibliographystyle{apsr}
 \bibliography{daildebates}

\end{spacing}

\end{document}